\PassOptionsToPackage{sort}{natbib}
\documentclass{article} 
\usepackage{iclr2027_conference,times}

\usepackage{hyperref}
\usepackage{url}

\usepackage[utf8]{inputenc} 
\usepackage[T1]{fontenc}    
\usepackage{graphicx}       
\usepackage{booktabs}       
\usepackage{adjustbox}      
\usepackage{caption}        
\usepackage{amsmath}        
\usepackage{amsfonts}       
\usepackage{microtype}      

\title{Evolving Layer-Specific Scalar Functions for Hardware-Aware Transformer Adaptation}

\author{%
  Kieran Carrigg$^1$ \\
  \texttt{kieran.carrigg@donders.ru.nl}
  \And
  Sigur de Vries$^1$ \\
  \texttt{sigur.devries@donders.ru.nl}
  \And
  Amirhossein Sadough$^1$ \\
  \texttt{amirhossein.sadough@donders.ru.nl}
  \And
  Marcel van Gerven$^1$ \\
  \texttt{marcel.vangerven@donders.ru.nl}
  \AND
  $^1$Department of Machine Learning and Neural Computing \\
  Donders Institute for Brain, Cognition, and Behaviour \\
  Thomas van Aquinostraat 4, 6525 GD Nijmegen, The Netherlands
}

\newif\ifarxiv
\arxivtrue 
\ifdefined\arxivbuild \arxivtrue \fi
\ifarxiv \iclrfinalcopy \fi

\begin{document}

\maketitle
\ifarxiv \lhead{} \renewcommand{\headrulewidth}{0pt} \fi

\begin{abstract}
   Vision Transformers (ViTs) achieve state-of-the-art performance on challenging vision tasks, but their deployment on edge devices is hindered by the computational complexity and global reduction bottleneck imposed by layer normalization. Recent methods attempt to bypass this by replacing normalization layers with hardware-friendly scalar approximations. However, these homogeneous replacements do not optimally fit to all layers' behaviour and rely on expensive model retraining. In this work, we propose a highly efficient, hardware-aware framework that utilizes genetic programming (GP) to evolve heterogeneous, layer-specific scalar functions directly from pre-trained network activations. Because each layer receives its own expression, the evolved functions adapt to the changing normalization behaviour across network depth. Coupled with a novel post-training re-alignment strategy, our approach eliminates the need to retrain models from scratch entirely. After only 20 epochs of re-alignment, our modified models recover $84.32\%$ Top-1 ImageNet-1K accuracy on ViT-B and $85.74\%$ on ViT-L. Across four pre-trained architectures (ViT-B, ViT-L, Swin-B, and ConvNeXt-B), our best variant stays within $0.8\%$ of the pre-trained model, and our expressions outperform the homogeneous Dynamic Tanh (DyT) baseline in 11 of 12 settings. By retaining near-baseline accuracy while removing the global reduction bottleneck, our expressions make normalization cheaper on edge hardware. In register-transfer level (RTL) simulations of two chips, they need about three times fewer compute cycles than LayerNorm and half of its off-chip reads on edge devices with small on-chip memory.
\end{abstract}

\section{Introduction}
Vision Transformers (ViTs) have emerged as state-of-the-art architectures across a wide range of computer vision tasks, achieving remarkable success in image classification, object detection, and semantic segmentation~\citep{dosovitskiy2020image, cao2025object, carion2020end, thisanke2023semantic}. By leveraging self-attention mechanisms to capture global context and long-range dependencies, ViTs consistently rival or outperform traditional convolutional neural networks, making them increasingly popular choices in modern computer vision tasks and research~\citep{raghu2021vision, liu2021swin}. However, their high computational complexity and reliance on operations such as layer normalization and Softmax severely hinder their deployment on resource-constrained edge devices. These non-linear components introduce complex operations that incur considerable latency and energy consumption, making real-time inference on mobile platforms, FPGAs, and ASICs notoriously difficult~\citep{xiao2025refining, sadeghi2024peano, zhao2025quark, marino2023me}. Because linear layers are now routinely optimized using low-precision integer quantization, the performance bottleneck in AI hardware has fundamentally shifted toward memory bandwidth and data movement~\citep{sun2025integer, xu2025hardware}. As a result, these complex non-linear operations disrupt optimized dataflows and consume a disproportionate amount of on-chip resources, occupying up to 60\% of memory usage and nearly 40\% of total execution time~\citep{chen2024vita, tabani2021improving}.

Among such non-linear operations, layer normalization (LayerNorm) is particularly costly relative to its arithmetic. Unlike batch normalization (BN) in convolutional networks, where the mean and variance can be calculated during training and folded into the preceding linear layers for inference, LayerNorm requires recalculating the mean and variance at the token level during inference~\citep{chen2025hardware, kim2025hardware}. This process involves a reduction operation across the feature dimension, which we define as the global reduction bottleneck. This operation creates a cross-feature data dependency that prevents layer fusion unless the whole feature vector is buffered on chip, and introduces inter-tile dependencies. Because the final normalization cannot occur until these statistics are computed sequentially, the system is forced into a multi-pass execution that repeatedly reads and writes intermediate data to off-chip memory. This saturates memory bandwidth and drives up both inference latency and energy consumption~\citep{sun2025integer, chen2025hardware, chen2024vita, lee2024q, yu2022nn, marino2023me, zhao2025quark, 10992712, 11130436,liu2023efficientvit,wang2023sole,dong2024eq}. Beyond this unavoidable data movement, the dynamic calculation of these token-dependent statistics imposes a high arithmetic stalling cost. Because the subsequent operations cannot proceed until the variance is computed and inverted, the primary computing units can be forced to idle, reducing end-to-end performance and system utilization on such hardware~\citep{xu2025hardware, xiao2025refining}.

To bypass these hardware and memory bottlenecks, recent works have explored various hardware-friendly approximations and normalization-free designs. Some approaches replace LayerNorm with BN to enable offline parameter folding~\citep{chen2025hardware}, while others substitute complex non-linearities with integer-based polynomial fitting or Newton-Raphson iterations~\citep{kim2025hardware, sun2025integer}. A notable advancement by~\citet{zhu2025transformers} proposes replacing LayerNorm with an element-wise operation called Dynamic Tanh (DyT), defined as $\text{DyT}(x)=\text{tanh}(\alpha x)$. This is based on the observation that LayerNorm mappings follow tanh-like, S-shaped curves. By replacing global statistics with a scalar mapping, DyT effectively eliminates the reduction bottleneck while achieving near-identical performance in ViTs. However, this method adopts a single, homogeneous scalar function for all LayerNorm layers using only a learnable scaling parameter $\alpha$ for flexibility. This overlooks the fact that normalization layers operate differently depending on network depth, transitioning from near-linear mappings in early blocks to the S-shaped curves observed in deeper ones~\citep{zhu2025transformers}. Additionally, this scalar replacement has primarily been evaluated through intensive from scratch training of 300 epochs or more~\citep{zhu2025transformers}. While this establishes the replacement's viability, the requirement for extensive retraining presents a significant barrier to the rapid hardware adaptation of existing pre-trained models. These factors highlight the need for a versatile framework capable of discovering layer-specific functions and adapting them efficiently to pre-trained architectures without the overhead of from-scratch training.

\begin{figure}[tp]
    \centering
    \includegraphics[width=\linewidth]{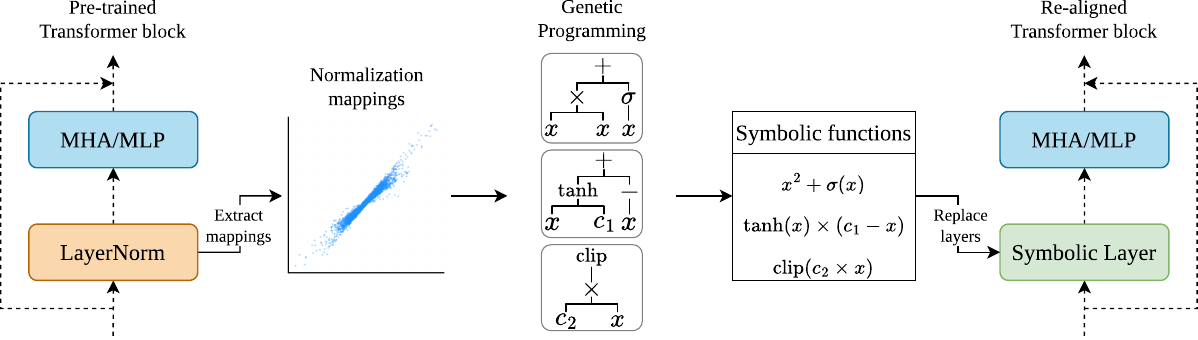}
    \caption{\textbf{Overview of the proposed symbolic discovery and re-alignment framework.} Normalization mappings are extracted from the LayerNorm operations within a pre-trained transformer block. Genetic programming uses these mappings to evolve layer-specific symbolic functions. We then replace the original LayerNorm operations with these discovered expressions, followed by a brief re-alignment phase to recover model performance.}
    \label{fig:framework_overview}
\end{figure}

We address these challenges with a two-fold hardware-aware adaptation strategy for pre-trained ViTs, summarized in Figure~\ref{fig:framework_overview}. First, we introduce a symbolic discovery framework using genetic programming (GP)~\citep{koza1994genetic} to evolve layer-specific scalar functions. While prior work~\citep{zhu2025transformers, chen2026stronger} demonstrated the viability of fixed scalar replacements, we automate this process via symbolic regression on the input-to-output mappings of pre-trained normalization layers. This allows our framework to discover heterogeneous expressions that capture the distinct normalization behaviours observed across the network. Second, we propose a novel, efficient post-training re-alignment strategy that recovers the model's feature space in a fraction of the standard training duration by leveraging existing LayerNorm weights and biases.

Our main contributions are summarized as follows:
\begin{itemize}
    \item We introduce a framework that utilizes GP to evolve heterogeneous, layer-specific scalar functions. These achieve significantly higher functional alignment than homogeneous baselines on all four architectures, reducing the mean squared error by a factor of $2$ to $4$ compared to DyT.
    \item We propose a computationally efficient post-training re-alignment strategy that leverages pre-trained weights to recover $84.32\%$ Top-1 ImageNet-1K accuracy on ViT-B (and $85.74\%$ on ViT-L) in only 20 epochs, completely avoiding the need for expensive retraining from scratch. Across four architectures (ViT-B, ViT-L, Swin-B, and ConvNeXt-B), our best variant stays within $0.8\%$ of the pre-trained model, and our expressions outperform the homogeneous DyT baseline in 11 of 12 settings.
    \item We remove the LayerNorm global reduction bottleneck, restoring an on-chip streaming dataflow. In register-transfer level (RTL) simulations of two chips, our expressions need about three times fewer compute cycles than LayerNorm and half of its off-chip reads on edge devices with small on-chip memory. Across all four architectures, they also need fewer FLOPs in total than both LayerNorm and DyT.
\end{itemize}

The remainder of this paper is organized as follows. Section~\ref{methods} details our extraction of normalization mappings, the GP symbolic discovery process, and the model re-alignment phase. Section~\ref{results} provides a comparative analysis of functional alignment, classification performance, and hardware efficiency. Finally, Section~\ref{discussion} discusses the implications and limitations of our findings. 

\section{Methods}
\label{methods}

In this section, we describe our framework for replacing LayerNorm with evolved symbolic functions. We conduct our experiments using pre-trained ViT-B and ViT-L models~\citep{dosovitskiy2020image} as our baseline architectures, leveraging their established feature representations to drive both our symbolic discovery and model re-alignment. To test whether our framework extends beyond a single architecture family, we include Swin-B~\citep{liu2021swin}, a hierarchical transformer with windowed attention, and ConvNeXt-B~\citep{liu2022convnet}, a convolutional network that also uses LayerNorm and was used to evaluate DyT by~\citet{zhu2025transformers}. Our implementation uses the official weights fine-tuned on ImageNet-1K~\citep{deng2009imagenet}, accessed via the \texttt{timm} library~\citep{rw2019timm}.
\ificlrfinal
    The source code is provided at \url{https://github.com/kierancarrigg/GP-LayerNorm}.
\else
    The anonymized source code is provided in the supplementary materials.
\fi

\subsection{Normalization mappings extraction}
\label{data_acq}
To extract the LayerNorm mappings for our symbolic search, we first consider its standard formulation:
\begin{equation}
    y = \frac{x - \mu}{\sqrt{\sigma^2 + \epsilon}} \odot w + b,
\end{equation}
where $\mu$ and $\sigma^2$ are the mean and variance over the channels of each token (or of each spatial position in ConvNeXt-B), $\epsilon$ is a small constant for numerical stability, and $w$ and $b$ are the per-channel learnable affine weights and biases. While the arithmetic operations themselves are relatively lightweight, the dynamic calculation of $\mu$ and $\sigma^2$ introduces the complex cross-feature reduction dependency that bottlenecks edge execution. To ensure our symbolic search discovers a pure, element-wise replacement for this specific reduction operation, we isolate the mapping prior to the affine transformation. Following an approach similar to~\citet{zhu2025transformers}, we propagate a single batch of 64 randomly drawn ImageNet-1K validation images through the pre-trained network (no labels are used). Given the captured layer output $y$, and the pre-trained weights $w$ and biases $b$, we compute the target pre-affine mapping $y_{pre}$ as:
\begin{equation}
    y_{pre}=\frac{y-b}{w+\epsilon}.
\end{equation}
Consistent with observations by~\citet{zhu2025transformers}, we find that normalization behaviour shifts significantly from near-linear mappings in earlier blocks to highly non-linear, S-shaped curves in deeper blocks (visualized across representative layers in Appendix~\ref{A:mappings}). To construct a representative dataset for symbolic regression, we randomly sample 50{,}000 data points across the feature and token dimensions per layer. This dataset is split into a 90/10 ratio over the points (not over images) for GP training and fitness validation, ensuring that the evolved solutions generalize well to unseen activations. Mappings extracted from training images instead give the same fit and accuracy (Appendix~\ref{A:calibration}). A complete visualization of these extracted mappings for all four architectures is provided in Appendix~\ref{A:equations}.

\subsection{Hardware-aware symbolic search}
\label{symbolic_search}
To discover optimal scalar replacements for each normalization layer, we utilize a symbolic discovery framework based on GP. We employ the \texttt{Kozax}~\citep{de2025kozax} library, a fast and flexible GP framework built in JAX, to conduct an independent evolutionary search for each normalization layer, run separately for ViT-B (25 layers), ViT-L (49 layers), Swin-B (53 layers), and ConvNeXt-B (41 layers). By fitting unique expressions to the individual mappings captured in Section~\ref{data_acq}, our approach discovers specialized replacements that accommodate the varying functional behaviours observed across the network depths.

In our GP implementation, candidate solutions are represented as parse trees with a maximum initial depth of 4 and a strict complexity limit of 20 nodes to ensure hardware friendliness. The operator set $\mathcal{O}$ consists of addition ($+$), multiplication ($\times$), the hyperbolic tangent ($\tanh$), the sigmoid function ($\sigma$), negation ($\text{neg}$), and value clipping to $\pm 5.0$ ($\text{clip}$). To optimize directly for hardware efficiency, we assign an estimated floating-point operations (FLOPs) cost to each operator, which serves as the complexity metric during the evolutionary search. The full derivation of these per-operator hardware costs is detailed in Appendix~\ref{A:Hardware-centric}. Finally, to improve search efficiency, constants within the trees are refined using gradient-based optimization for 10 steps per evaluation.

The primary objective of the search is to minimize the error between the evolved function $f(x)$ and the target mapping $y_{pre}$. However, our preliminary experiments revealed that optimizing for mean squared error (MSE) alone sometimes favoured near-linear functions that accurately captured the central distribution but exhibited limitless growth at the activation tails. When integrated into the ViT-B architecture, these unconstrained functions caused activation explosion, with NaN losses within the first two epochs of fine-tuning (Appendix~\ref{A:ablations}). To address this, we developed a composite fitness function $\mathcal{F}$ designed to enforce numerical stability via a "pull-to-zero" penalty:
\begin{equation}
    \mathcal{F}=\text{MSE}(y_{pre}, f(x))+\gamma \cdot \tfrac{1}{2} \sum_{x \in \{-\delta, \delta\}}f(x)^2,
\end{equation}
where $\gamma=0.005$ is a weighting factor determined through empirical tuning to balance fitting accuracy with model stability. The anchor $\delta$ is dynamically set to $2 \cdot \max|x_{train}|$ for each layer, forcing the GP to favour expressions that revert toward zero at out-of-distribution extremes. This constraint effectively mimics the regularizing properties of standard LayerNorm, ensuring that the evolved scalar functions behave similarly during the model re-alignment.

For each search, the framework utilizes the NSGA-II algorithm~\citep{deb2002fast} to perform multi-objective optimization, treating both the composite fitness $\mathcal{F}$ and functional complexity (estimated FLOPs) as competing objectives. Without this cost objective, the search reaches about the same accuracy with expressions that cost about $15\times$ more (Appendix~\ref{A:ablations}). We set a population size of 500 evolved over 50 generations. To account for the stochastic nature of GP, we perform five independent evolutionary searches for every normalization layer using varying random seeds. For each layer-seed combination, we maintain a complete Pareto front of solutions. To identify the optimal balance between functional fitness and computational cost, we apply the Kneedle algorithm~\citep{satopaa2011finding} to locate the ``knee'' point of the Pareto front. Before this step, we discard candidates that grow faster than linearly or are not monotonic, as both caused unstable re-alignment in preliminary experiments. This process yields five distinct sets of symbolic replacements for the normalization layers in the network, providing the foundation for the model re-alignment phase detailed in Section~\ref{model_realign}.

\subsection{Model re-alignment}
\label{model_realign}
The final stage of our framework involves restoring the classification performance of the pre-trained models. We replace all LayerNorm layers with corresponding element-wise modules: either the symbolic solutions obtained with our GP framework (see Section~\ref{symbolic_search}), DyT layers, or Dynamic erf (Derf) layers~\citep{chen2026stronger}, a follow-up to DyT that replaces $\tanh$ with the error function and adds a learnable shift $s$, defined as $\text{Derf}(x)=\text{erf}(\alpha x + s)$. To ensure minimal disruption to the network's established feature representations, these custom modules inherit the pre-trained affine weights ($w$) and biases ($b$) from the original layers. Preserving these parameters is essential for functional consistency, as our GP expressions operate as fixed approximations of the model's pre-affine mappings. For the DyT and Derf modules, we initialize the learnable $\alpha$ parameter at 0.5 to maintain consistency with~\citet{zhu2025transformers}, and Derf's shift $s$ at 0.

We evaluate each method on the ImageNet-1K training set using three fine-tuning variants: -A (affine-only), -F (full fine-tuning), and -D (full fine-tuning with distillation). For the affine-only variants (GP-A, DyT-A, and Derf-A), the backbone is frozen, optimizing only the learnable affine weights and biases ($w, b$). For the full fine-tuning variants (GP-F, DyT-F, and Derf-F), the entire network is unfrozen to allow global feature re-alignment. Finally, the distillation variants (GP-D, DyT-D, and Derf-D) also unfreeze the entire network, employing logit-based distillation~\citep{hinton2015distilling} where the student model minimizes the Kullback-Leibler (KL) divergence against its own pre-trained, LayerNorm-based teacher alongside standard cross-entropy loss. As a reference, we report the accuracy of the original pre-trained model (LN).

To ensure statistical robustness against the stochastic nature of both symbolic search and re-alignment, we report results averaged over five independent runs. GP variants utilize the five distinct sets of evolved functions (Section~\ref{symbolic_search}), while DyT and Derf vary the training initialization seed. All models are trained for 20 epochs using AdamW without warmup, applying a cosine learning rate schedule for most variants; the complete per-variant, per-architecture schedule is detailed in Appendix~\ref{A:hyperparams}. We performed independent grid searches per variant to optimize the peak learning rate, weight decay, and stochastic depth. For Swin-B and ConvNeXt-B, we searched only the learning rate and kept the weight decay and stochastic depth from the ViT-B experiments. The general experimental configuration is detailed in Appendix~\ref{A:exp_settings}. Notably, we observed a significant disparity in training stability at ViT-B and ViT-L: GP trained stably using a single global learning rate, whereas the DyT-F and DyT-D variants were highly prone to divergence, necessitating differential learning rates across the backbone, affine, and $\alpha$ parameters. We kept these differential learning rates for DyT on Swin-B and ConvNeXt-B and used them for Derf on all four architectures, while GP again used a single learning rate. For the distillation variants, the total loss $\mathcal{L}$ is the weighted sum of the standard cross-entropy loss ($\mathcal{L}_{CE}$) and the KL-divergence ($\mathcal{L}_{KL}$):
\begin{equation}
    \mathcal{L}=(1-\lambda)\mathcal{L}_{CE}+\lambda \tau^2\mathcal{L}_{KL},
\end{equation}
where the loss balancing coefficient $\lambda=0.5$ and temperature $\tau=4.0$ were empirically optimized in preliminary experiments.

\section{Results}
\label{results}

\subsection{Symbolic alignment}

\begin{figure}[tbp]
    \centering
    \includegraphics[width=\textwidth]{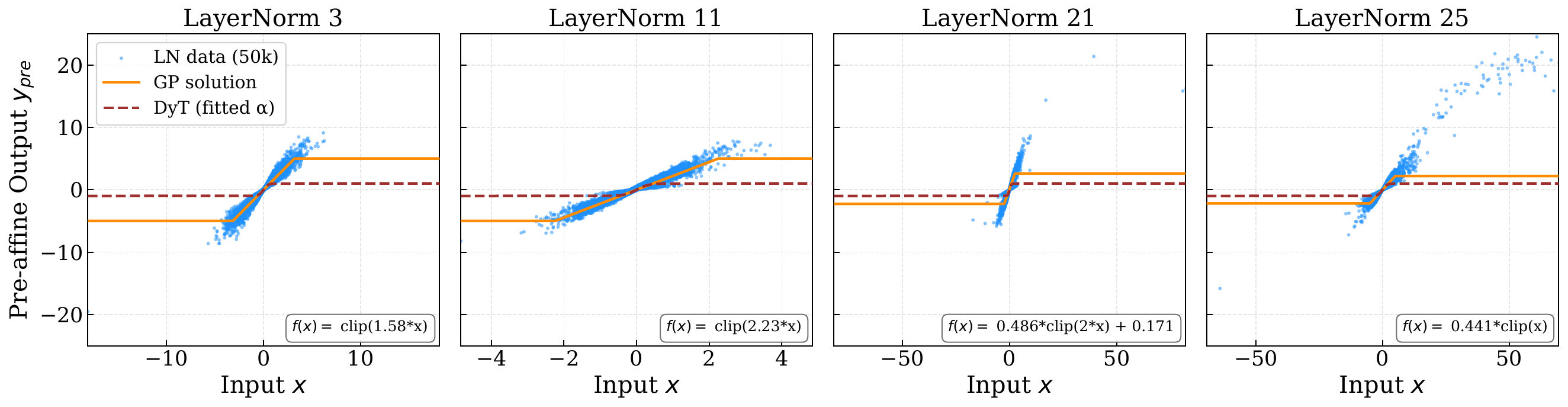}
    \caption{\textbf{Functional alignment of discovered symbolic expressions on ViT-B.} The evolved GP solutions (orange lines) and the optimized DyT baseline (dashed red lines, utilizing a least-squares optimized $\alpha$) are overlaid onto the 50{,}000-point LayerNorm mappings (blue scatter). The GP framework successfully discovers symbolic functions that match both the shape and amplitude of the extracted LayerNorm mappings across all network depths. The explicitly discovered mathematical functions for each layer are displayed in the text boxes.}
    \label{fig:gp_alignment}
\end{figure}

We first evaluate the capacity of our GP framework to capture the diverse functional behaviours of pre-trained LayerNorm layers. Figure~\ref{fig:gp_alignment} visualizes the discovered symbolic expressions overlaid on the target mappings for four representative layers across ViT-B's network depth. As observed in Section~\ref{data_acq}, the normalization behaviour shifts significantly from near-linear mappings in earlier blocks to highly non-linear, S-shaped curves in deeper blocks. Our framework successfully discovers heterogeneous expressions that adapt to these varying behaviours. For instance, in earlier layers, the discovered solutions are nearly linear, often using the \texttt{clip} operator to manage activation outliers and mimic the regularizing properties of LayerNorm. For the pronounced S-curves of deeper blocks, the GP scales and shifts those same \texttt{clip} operators, constructing piecewise-linear approximations at low computational cost. This choice does not come from the cost objective alone: without it, $21$ of the $25$ ViT-B expressions already contain \texttt{clip}, and the cost objective mainly removes the remaining $\tanh$ and $\sigma$ terms, at the same accuracy (Appendix~\ref{A:ablations}). A comprehensive visualization of the LayerNorm mappings and symbolic expressions for all four architectures is provided in Appendix~\ref{A:equations}.

\begin{table}[bp]
  \small
  \centering
  \caption{\textbf{Functional alignment quality.} Mean squared error (MSE, $\downarrow$) of GP-discovered expressions, Dynamic Tanh (DyT), and Dynamic erf (Derf) against the 50{,}000-point LayerNorm mappings of four pre-trained architectures. GP results represent the mean $\pm$ std across five independent seeds. Both baselines use per-layer least-squares optimized parameters: $\alpha$ for DyT, and $\alpha$ and shift $s$ for Derf.}
  \label{tab:alignment_quality}
  \begingroup
  \begin{tabular}{lcccc}
    \toprule
    \textbf{Method} & \textbf{ViT-B} & \textbf{ViT-L} & \textbf{Swin-B} & \textbf{ConvNeXt-B} \\
    \midrule
    DyT (Optimized $\alpha$) & $0.300$ & $0.244$ & $0.403$ & $0.371$ \\
    Derf (Optimized $\alpha$, $s$) & $0.275$ & $0.240$ & $0.399$ & $0.365$ \\
    \textbf{GP (Ours)} & $\mathbf{0.100 \pm 0.004}$ & $\mathbf{0.073 \pm 0.001}$ & $\mathbf{0.183 \pm 0.002}$ & $\mathbf{0.093 \pm 0.002}$ \\
    \bottomrule
  \end{tabular}
  \endgroup
\end{table}

To quantify the quality of this alignment, we report the mean squared error (MSE) in Table~\ref{tab:alignment_quality}, averaged over all normalization layers of each architecture. We compare our results against DyT and Derf baselines, where $\alpha$ (and Derf's shift $s$) was independently optimized for every layer using least-squares regression. Our GP solutions achieve a substantially better alignment with the original LayerNorm mappings on all four architectures, with an MSE $2$ to $4$ times lower than both baselines. Because the pre-affine LayerNorm outputs have unit variance, the MSE is close to the fraction of the variance that is left unexplained: our expressions explain $82$--$93\%$ of the variance across the four architectures, against $60$--$76\%$ for DyT and Derf. Swin-B is the hardest to fit for all three methods, yet GP still halves the error of both baselines. Much of this gap comes from amplitude. DyT and Derf are bounded to $[-1,1]$, whereas the LayerNorm mappings are not (Figure~\ref{fig:gp_alignment}). Our expressions, in contrast, can scale their output to match each layer. The standard deviation across the five seeds is small (at most $\pm 0.004$), so the search finds expressions of similar quality regardless of its random seed. However, a closer fit does not always give a better model (Appendix~\ref{A:clip}). We therefore compare the methods on the accuracy they recover after replacement (Section~\ref{recovery}).

\subsection{Performance recovery}
\label{recovery}
To evaluate the effectiveness of our symbolic replacements, we measure the performance recovery during a 20 epoch re-alignment phase, where the model is fine-tuned after replacing its normalization layers. This process reveals a significant disparity in how different functional approximations impact the pre-trained feature space. The initial replacement of LayerNorm with element-wise functions causes a distinct "shock" to the model's performance: at epoch 0, before any fine-tuning, all models with replaced LayerNorm layers show a significant drop in performance from the pre-trained baseline. However, the difference in this performance drop already shows a major advantage of our heterogeneous GP solutions compared to the homogeneous DyT approximations. Before any tuning has occurred, the GP variants already maintain a Top-1 accuracy of approximately $51.4\%$ at ViT-B and $22.3\%$ at ViT-L, suggesting that our discovered symbolic functions successfully preserve a meaningful portion of the pre-trained feature representations, though this retention is markedly weaker at the larger scale. In contrast, the homogeneous DyT approximations cause the model to break entirely, collapsing to chance level ($0.10\%$ and $0.11\%$, respectively) at initialization, as does Derf on all four architectures (at most $0.15\%$). At ConvNeXt-B, GP retains $11.9\%$ while DyT again collapses ($0.08\%$). At Swin-B, however, both methods start close to chance ($0.25\%$ for GP and $0.10\%$ for DyT), yet GP still recovers faster (after one epoch, $81.7\%$ for GP-D versus $75.2\%$ for DyT-D) and reaches a higher final accuracy in every fine-tuning variant, indicating that the advantage of our symbolic functions is not limited to the accuracy retained at replacement.

\begin{figure}[tbp]
    \centering
    \includegraphics[width=\textwidth]{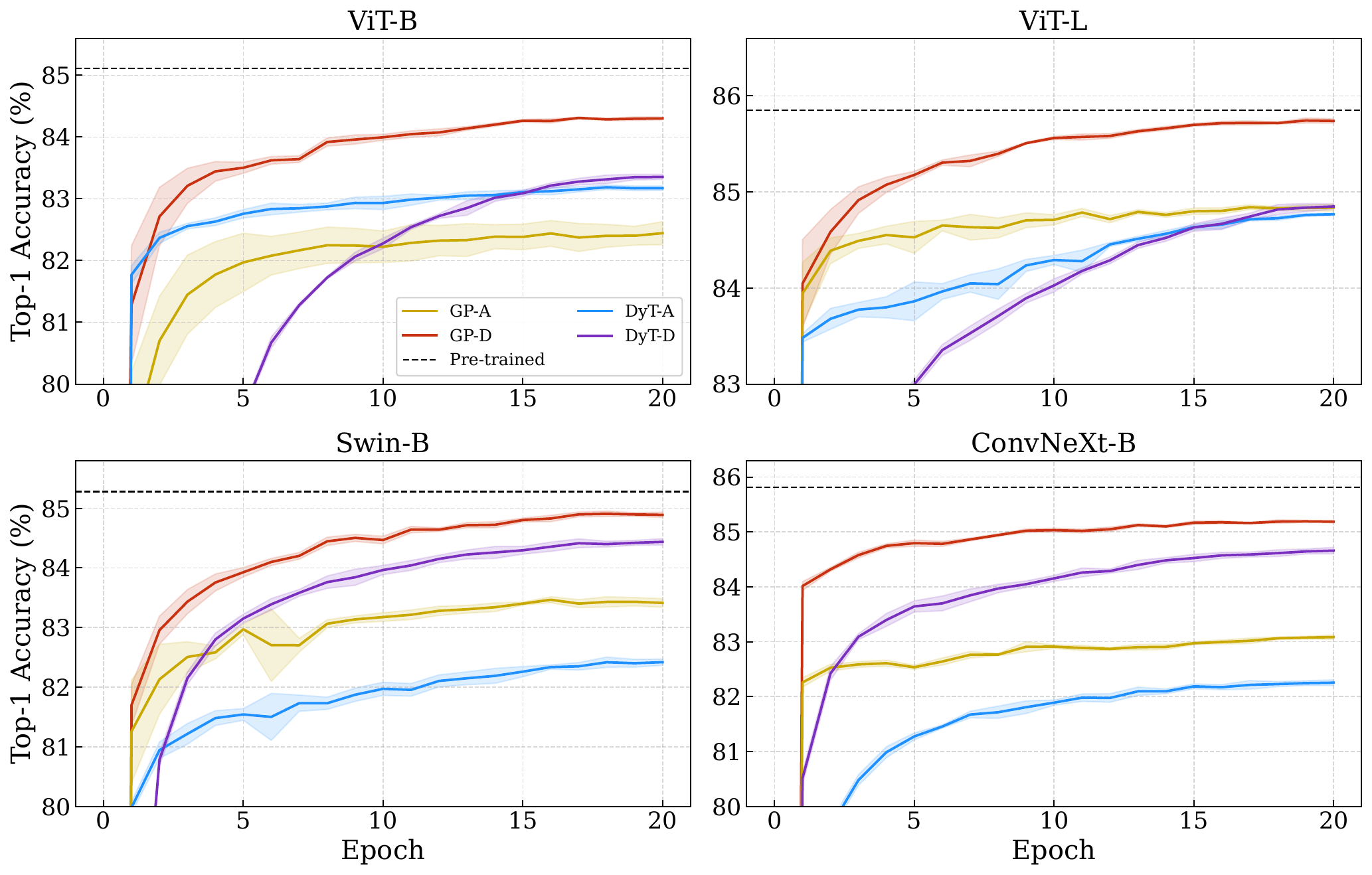}
    \caption{\textbf{ImageNet-1K validation performance recovery dynamics.} Fine-tuning trajectories of GP-A, DyT-A, GP-D, and DyT-D at ViT-B (top left), ViT-L (top right), Swin-B (bottom left), and ConvNeXt-B (bottom right). Shaded areas represent $\pm 1$ standard deviation across five independent seeds. Black dashed lines show the accuracy of the original pre-trained model ($85.11\%$, $85.85\%$, $85.28\%$, and $85.81\%$, respectively). The full fine-tuning variants (GP-F and DyT-F) and Derf, whose trajectories closely follow those of DyT, are shown in Appendix~\ref{A:finetune}.}
    \label{fig:accuracy_curves}
\end{figure}

The remainder of the performance recovery process demonstrates both our heterogeneous layer-specific approach and our efficient post-training strategy. As Figure~\ref{fig:accuracy_curves} shows, GP-D is the highest curve on all four architectures from the second epoch onwards, whereas the affine-only variants (GP-A and DyT-A) level off at lower accuracies. On both ViTs, DyT-D converges much more slowly than GP-D. On Swin-B and ConvNeXt-B it recovers much faster than on the ViTs, but stays below GP-D throughout. The full fine-tuning variants (GP-F and DyT-F) follow the same pattern and are analysed in Appendix~\ref{A:finetune}.

The peak quantitative results for this re-alignment are summarized in Table~\ref{tab:main_results}. Our best-performing variant, GP-D, reaches a peak Top-1 accuracy of $84.32\%$ at ViT-B and $85.74\%$ at ViT-L, and stays within $0.8\%$ of the original pre-trained model on all four architectures ($0.79\%$, $0.11\%$, $0.36\%$, and $0.60\%$ below it at ViT-B, ViT-L, Swin-B, and ConvNeXt-B, respectively). The one setting where GP trails is affine-only fine-tuning at ViT-B, where both DyT and Derf are ahead; in every other comparison across the four architectures, our GP solutions exceed both baselines. This also holds under distribution shift: on ImageNet-V2~\citep{recht2019imagenet}, ImageNet-Sketch~\citep{wang2019learning}, ImageNet-R~\citep{hendrycks2021many}, and ImageNet-A~\citep{hendrycks2021natural}, GP-F and GP-D outperform DyT-F and DyT-D on every test set at all four architectures (Appendix~\ref{A:robustness}). Furthermore, these results validate our broader post-training strategy; both our DyT-F and DyT-D variants surpass the accuracies reported in the original study~\citep{zhu2025transformers} at ViT-B ($82.5\%$), ViT-L ($83.6\%$), and ConvNeXt-B ($83.7\%$), which required training from scratch for over 300 epochs. This demonstrates that a short re-alignment phase is a viable and efficient alternative to training normalization-free architectures from initialization.

\begin{table}[tbp]
  \small
  \centering
  \caption{\textbf{Main fine-tuning results on the ImageNet-1K validation set.} Peak Top-1 accuracy (\%) of our evolved symbolic normalizations (GP) compared against standard LayerNorm, Dynamic Tanh (DyT), and Dynamic erf (Derf) baselines on four pre-trained architectures, for affine-only fine-tuning (-A), full fine-tuning (-F), and knowledge distillation (-D). All variants are fine-tuned for 20 epochs and evaluated across 5 independent runs (mean $\pm$ standard deviation). The LN row is the original pre-trained model without fine-tuning.}
  \label{tab:main_results}
  \begin{tabular}{lcccc}
    \toprule
    \textbf{Method} & \textbf{ViT-B} & \textbf{ViT-L} & \textbf{Swin-B} & \textbf{ConvNeXt-B} \\
    \midrule
    LN (Pre-trained) & $85.11$ & $85.85$ & $85.28$ & $85.81$ \\
    \midrule
    DyT-A & $\mathbf{83.19 \pm 0.03}$ & $84.77 \pm 0.01$ & $82.43 \pm 0.07$ & $82.27 \pm 0.04$ \\
    Derf-A & $83.14 \pm 0.02$ & $84.59 \pm 0.03$ & $82.14 \pm 0.07$ & $82.04 \pm 0.02$ \\
    GP-A (Ours) & $82.48 \pm 0.20$ & $\mathbf{84.85 \pm 0.02}$ & $\mathbf{83.48 \pm 0.06}$ & $\mathbf{83.09 \pm 0.03}$ \\
    \midrule
    DyT-F & $82.81 \pm 0.02$ & $84.23 \pm 0.05$ & $83.40 \pm 0.09$ & $83.93 \pm 0.05$ \\
    Derf-F & $82.36 \pm 0.19$ & $84.37 \pm 0.09$ & $83.45 \pm 0.08$ & $83.91 \pm 0.07$ \\
    GP-F (Ours) & $\mathbf{84.07 \pm 0.06}$ & $\mathbf{85.19 \pm 0.02}$ & $\mathbf{84.21 \pm 0.05}$ & $\mathbf{84.85 \pm 0.03}$ \\
    \midrule
    DyT-D & $83.36 \pm 0.04$ & $84.85 \pm 0.03$ & $84.45 \pm 0.05$ & $84.66 \pm 0.06$ \\
    Derf-D & $83.55 \pm 0.08$ & $84.93 \pm 0.04$ & $84.51 \pm 0.10$ & $84.65 \pm 0.08$ \\
    GP-D (Ours) & $\mathbf{84.32 \pm 0.01}$ & $\mathbf{85.74 \pm 0.02}$ & $\mathbf{84.92 \pm 0.04}$ & $\mathbf{85.21 \pm 0.01}$ \\
    \bottomrule
  \end{tabular}
\end{table}

\subsection{Hardware efficiency}
\label{sec:complexity}
\begin{figure}[t]
    \centering
    \includegraphics[width=\textwidth]{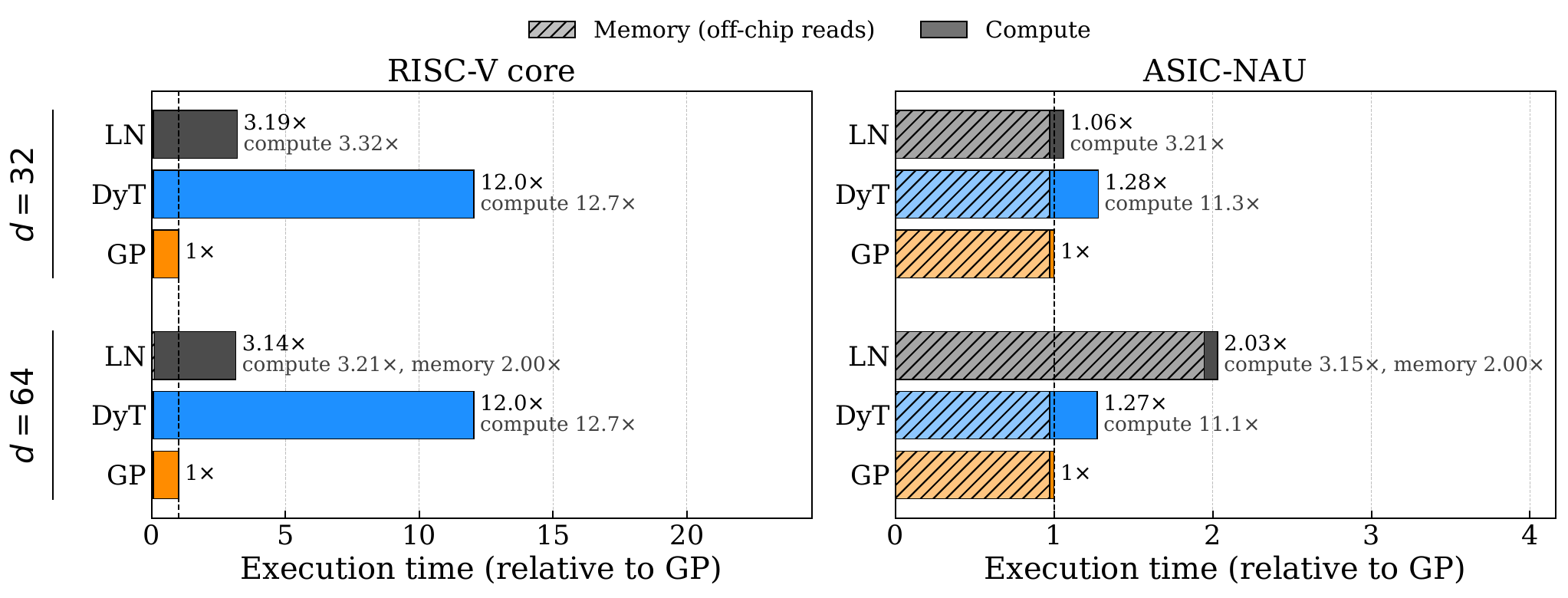}
\caption{\textbf{Execution time per normalization layer from RTL simulations of ViT-B, relative to GP.} Each bar is split into the memory phase (hatched, off-chip reads) and the compute phase (solid). The dashed line marks GP. Labels give the total and compute time relative to GP at the same $d$, and the memory time where it differs from GP's (LayerNorm at $d = 64$, which reads every input twice). With a buffer of $M = 32$ values, the vector fits at $d = 32$ and spans two tiles at $d = 64$. Absolute cycle counts are given in Appendix~\ref{A:rtl}. \textbf{(Left)} RISC-V core, where the computation takes most of the time, so the total ratios are close to the compute ratios. \textbf{(Right)} ASIC-NAU, where a programmable nonlinear unit runs the normalization and the off-chip reads take most of the time, so the total ratios are much smaller than the compute ratios.}
\label{fig:rtl_cycles}
\end{figure}

Edge AI accelerators have limited on-chip memory, so weights and intermediate activations are usually kept in off-chip memory and brought on chip in tiles~\citep{7738524, 11343916, 10181988, 10558587}. Off-chip transfers then account for a large part of the latency and energy~\citep{10992712, 11130436}, in one recent edge design up to $52\%$ of the total energy~\citep{10558587}. Dataflows that keep tiles on chip and avoid intermediate write-backs are therefore strongly preferred. LayerNorm's global reduction prevents this streaming execution by forcing off-chip round-trips, so it can add a significant latency even though it needs few operations~\citep{10558587}. These bandwidth-bound devices are the setting we target in this section. Let $M$ be the number of values the buffer of the normalization unit holds and $d$ the length of the vector to be normalized, so that the vector is brought on chip in $\lceil d/M \rceil$ tiles. LayerNorm needs the mean and variance of the whole vector before it can normalize any element, so when $d > M$ it reads every tile twice from off-chip memory (once to compute the statistics and once to normalize). It can avoid the second read by keeping the vector on chip, but this needs a buffer of at least $d$ values and a loop order that fuses LayerNorm with the preceding layer~\citep{10558587}. DyT and our GP expressions are elementwise, so they read each tile once for any $d$.

We measure both cases with register-transfer level (RTL) simulations of two chips built on Caravel~\citep{caravel}, an open-source SoC in the SkyWater 130\,nm process that has been proven in silicon. The first runs all three methods on the SoC's VexRiscv RV32I core, which has no floating-point unit. Its 1.5\,KB of SRAM could hold at most 384 FP32 values, well below common hidden sizes of 768 to 4096, so on such a chip vectors must be tiled. The second chip, ASIC-NAU, adds a programmable nonlinear accelerator unit (NAU) to the same SoC. The NAU holds the buffer and runs the normalization, and the core only moves the inputs from the off-chip flash memory into this buffer. Because every run simulates the whole chip, we keep the sizes small: a buffer of $M = 32$ values and vectors of $d = 32$ (one tile) and $d = 64$ (two tiles). Larger $d$ only adds more tiles of the same pattern. For GP we use the ViT-B expressions of seed 1, averaged over its $25$ layers, and DyT uses the exact $\tanh$. We split each run into a memory phase, reading the inputs from the flash, and a compute phase, the normalization itself. The energy of each phase is estimated with OpenSTA from the chip's signoff netlist and the switching activity of the simulation, and excludes the external flash memory. Appendix~\ref{A:rtl} gives the full setup and results.

Figure~\ref{fig:rtl_cycles} shows the execution time per layer, relative to GP. On the RISC-V core the computation dominates, and LayerNorm takes about $3.2\times$ the cycles of GP and DyT about $12\times$ at both sizes, with almost the same ratios in energy ($3.1\times$ and $12\times$). On ASIC-NAU it is the other way around: the computation is cheap and the off-chip reads dominate. When the vector fits in the buffer ($d = 32$), all three methods read their inputs once and differ only in the short compute phase, which for LayerNorm includes its two passes over the buffered elements. LayerNorm then takes $1.06\times$ and DyT $1.28\times$ the cycles of GP. When the vector spans two tiles ($d = 64$), LayerNorm reads every input twice and takes $2.03\times$ the cycles and $2.06\times$ the energy of GP, while DyT stays at $1.27\times$ and $1.38\times$. For $d = kM$, LayerNorm reads $2k$ tiles against $k$ for GP. This means that its memory cost stays twice that of GP at realistic hidden sizes, where the vector no longer fits in a small buffer. These measurements are per normalization layer. How much they shorten a full forward pass depends on the share of normalization in the model's runtime on a given device. They are also conservative. Each run has the flash interface to itself, whereas in a real system the memory bandwidth is shared with other components, and the excluded flash energy would add to every off-chip read. Both would increase the cost of LayerNorm's second read.

Beyond these measurements, an analytical count covers the full models. Counting each arithmetic operation as one FLOP (Appendix~\ref{A:Hardware-centric}), LayerNorm needs $5d + 3$ FLOPs per token and DyT $24d$, both excluding the shared per-channel affine transform $(w, b)$. The cost of our GP expressions varies layer by layer (Appendix~\ref{app:memory}, Figure~\ref{fig:complexity}) and follows the shape of the mapping. At ViT-B (seed 1), $21$ of the $25$ layers need a single operation per element, and only a few deeper layers need two or three operations. In total, our expressions need fewer FLOPs than both LayerNorm and DyT on all four architectures, at ViT-B roughly $3$--$4\times$ fewer than LayerNorm across the five seeds.

\section{Discussion}
\label{discussion}
The global reduction bottleneck makes LayerNorm costly on edge devices relative to its arithmetic. While high-bandwidth GPUs can efficiently absorb the cost of this cross-feature data dependency, modern edge accelerators are primarily bottlenecked by memory bandwidth rather than arithmetic throughput~\citep{10454330,10904793}. We address this bottleneck by using GP to evolve heterogeneous, layer-specific scalar functions directly from pre-trained network activations. By breaking the cross-feature dependency that ties LayerNorm to its global reduction, our approach enables optimizations such as token-wise tiling and layer fusion. In our RTL simulations, our expressions need fewer off-chip reads and fewer compute cycles than LayerNorm, and they recover higher accuracy than homogeneous scalar replacements after fine-tuning.

Compared to homogeneous scalar replacements~\citep{zhu2025transformers, chen2026stronger}, our evolved expressions achieve a closer functional alignment with the LayerNorm mappings of all four architectures (see Table~\ref{tab:alignment_quality}). Because every layer receives its own function, the expressions follow the shift from near-linear mappings in earlier blocks to S-shaped mappings in deeper blocks (Figure~\ref{fig:gp_alignment}), which a single $\tanh$ shape cannot. This alignment helps to retain accuracy at replacement on three of the four architectures, where homogeneous functions collapse to chance level (Section~\ref{recovery}). At Swin-B, our expressions also start close to chance, yet our full fine-tuning (GP-F) and distillation (GP-D) variants still reach higher final accuracies than their DyT counterparts, as on the other three architectures. This suggests that the accuracy at replacement understates how much of the pre-trained representations our expressions preserve. When the backbone is frozen, our affine-only variant (GP-A) also outperforms DyT-A at ViT-L, Swin-B, and ConvNeXt-B, with ViT-B as the only exception.

Our search also points to a common shape. DyT and Derf start from an assumed S-shaped function ($\tanh$ and erf). Although $\tanh$ and $\sigma$ were also available to our search, it selected the \texttt{clip} operator in every layer of all four architectures, most often as a scaled clip alone, which costs a single multiplication per element (Appendix~\ref{A:equations}). This form covers $90\%$ of the layers at ViT-B, where we test Clip-D below, and $60$ to $76\%$ at the other three architectures, where most of the remaining layers add a linear term, $a \cdot \mathrm{clip}(bx) + dx$. The clip shape is therefore a finding of the search itself, not an assumption we made beforehand. Once found, the shape can be reused directly. Fitting a single scaled clip to every layer of ViT-B (Clip-D, Appendix~\ref{A:clip}) reaches $84.22\%$ Top-1, which closes $0.86$ of the $0.96$ points between DyT-D and GP-D, while the per-layer search still adds a small but consistent gain on top. Besides layer-specific expressions, our method can therefore also bring out cheap functional forms such as this one, which can then be studied and reused on their own.

Because our symbolic expressions recover near-baseline accuracy after fine-tuning, we can fully realize the hardware benefits of breaking the global reduction bottleneck. Prior element-wise substitutions trade an increase in on-chip compute for reduced data movement. In our simulations, DyT halves the off-chip reads of LayerNorm once the vector is tiled, but needs $3.5$ to $4$ times its compute cycles on both chips (Appendix~\ref{A:rtl}, Table~\ref{tab:rtl}). Our layer-specific expressions reduce both, with the memory traffic of DyT and about three times fewer compute cycles than LayerNorm. The analytical count extends this to all four architectures, where our GP solutions require fewer operations in total than standard LayerNorm, DyT, and Derf (Appendix~\ref{app:gp-flops}). At the same time, our best GP variant reaches higher final accuracy than the best DyT and Derf variants on all four architectures. The heterogeneity of these expressions does not require unique hardware circuits for every layer. Each expression is a composition of a small set of primitives, so a single reconfigurable activation unit can evaluate all of them on a shared datapath. Existing designs such as COMPACT~\citep{COMPACT}, Flex-PE~\citep{FlexPE}, and LoRA~\citep{LORA} already follow this principle: a new expression changes only the operation sequence or the stored coefficients, not the hardware. Our simulations follow the same pattern, as both chips evaluate LayerNorm, DyT, and every GP layer on the same hardware and differ only in the program they run. Sharing the datapath also lets simpler expressions skip expensive transcendental steps, which lowers their latency and energy.

Alongside these results, we note several limitations and avenues for future work. First, a closer per-layer alignment does not always give a better model. At ViT-L, the stronger per-layer alignment does not yield higher accuracy at replacement than at ViT-B, and at Swin-B per-layer fitting alone retains almost no accuracy at replacement. Future work could therefore evaluate model accuracy directly within the GP evolutionary loop. Despite heavy computational costs, this directly optimizes the true objective and captures cross-layer interactions. Second, our hardware results come from RTL simulations of two small chips with short vectors ($d \le 64$), for one model and one set of expressions, measured per normalization layer. Measuring a full model at realistic hidden sizes on a deployed accelerator would show how much of this gain carries over end to end.

While our validation spans three architecture families (plain and hierarchical transformers, and a convolutional network), it covers a single dataset and task (ImageNet-1K classification).~\citet{zhu2025transformers} demonstrated that homogeneous scalar replacements are viable across a range of tasks and domains, including language models. Whether our heterogeneous GP framework generalizes in the same way, for instance to object detection, segmentation, or NLP, remains an open question. Beyond generalization, our layer-specific approach leaves room for further hardware savings. Because we evolve a unique equation for every layer, the deployed model must store one instruction set per normalization layer. Clip-D (Appendix~\ref{A:clip}), which reuses the shape found by our search, shows that this is not always needed: one functional form shared by all layers of ViT-B, with only its two constants fitted per layer, comes within $0.10$ points of GP-D. Future work could build this sharing into the search, for instance by constraining layers with similar normalization mappings to reuse the same symbolic sub-trees, or by searching for a single form for all layers directly. This would reduce the instruction memory required on the target edge device, at a small cost in accuracy.

\subsection*{AI use statement}

In this work, we used generative AI tools to implement methods, and to clean and reformat experimental data. The methods themselves, including the symbolic search formulation, the composite fitness function, and the post-training re-alignment procedure, were conceived by the authors; generative AI assistance was limited to their implementation in code. We have not used generative AI tools to develop theoretical models or conceptual frameworks, to formulate mathematical claims or assist in their derivation, to formulate or refine hypotheses, to design or provide feedback on research methodology or experiments, or to interpret results, and the generation of synthetic data sets and translation are not applicable to this work.

Additionally, we used generative AI tools to create and edit software code, to create and modify figures, to draft and edit parts of this paper for readability and formatting, to summarize existing literature, and to format references.

We have reviewed all AI-assisted work. All AI-implemented code was reviewed and tested by the authors against the intended method specification, all reported numerical results were verified against the original experiment logs, and all AI-assisted text was reviewed and revised by the authors. We take responsibility for the final content of this work, including text, claims or artifacts produced with the aid of generative AI.

\subsection*{Ethics statement}

By reducing the cost of normalization on edge hardware, our framework makes it easier to run highly capable vision models directly on local, resource-constrained edge devices. While this local processing improves latency, reduces cloud reliance, and lowers the energy footprint of AI systems, it inherently introduces dual-use risks. Specifically, unconstrained edge vision could facilitate pervasive surveillance systems, raising privacy concerns, or be adapted for autonomous military applications. We encourage the research community to prioritize transparent, ethical frameworks when deploying hardware-aware vision models in real-world environments.

\subsection*{Reproducibility statement}

We have taken several steps to ensure that our results can be reproduced. The complete framework, covering normalization mapping extraction, the GP search, and the model re-alignment procedure, is described in Section~\ref{methods}, with the general experimental configuration given in Appendix~\ref{A:exp_settings} and the per-variant, per-architecture hyperparameters in Appendix~\ref{A:hyperparams}. Appendix~\ref{A:equations} lists the exact symbolic expressions discovered for every normalization layer of all four architectures, allowing the modified models to be constructed directly without repeating the evolutionary search. All reported results are averaged over five independent runs, with the resulting variability shown as error shading in Figure~\ref{fig:accuracy_curves} and as standard deviations in Table~\ref{tab:alignment_quality} and Table~\ref{tab:main_results}. Finally, \ificlrfinal the code for the GP search and model re-alignment is available at \url{https://github.com/kierancarrigg/GP-LayerNorm}.\else we provide an anonymized archive of all code and scripts required to reproduce our main experimental results, together with the instructions and settings used, as supplementary material.\fi


\ificlrfinal
\subsubsection*{Acknowledgments}
    This publication is part of the project ROBUST: Trustworthy AI-based Systems for Sustainable Growth with project number KICH3.LTP.20.006, which is (partly) financed by the Dutch Research Council (NWO), ASMPT, and the Dutch Ministry of Economic Affairs and Climate Policy (EZK) under the program LTP KIC 2020-2023. All content represents the opinion of the authors, which is not necessarily shared or endorsed by their respective employers and/or sponsors.
    This work used the Dutch national e-infrastructure with the support of the SURF Cooperative using grant no. EINF-21110
\fi

\bibliography{references}

\begin{thebibliography}{53}
\providecommand{\natexlab}[1]{#1}
\providecommand{\url}[1]{\texttt{#1}}
\expandafter\ifx\csname urlstyle\endcsname\relax
  \providecommand{\doi}[1]{doi: #1}\else
  \providecommand{\doi}{doi: \begingroup \urlstyle{rm}\Url}\fi

\bibitem[Abramowitz \& Stegun(1964)Abramowitz and Stegun]{abramowitz1964handbook}
Milton Abramowitz and Irene~A. Stegun.
\newblock \emph{Handbook of Mathematical Functions with Formulas, Graphs, and Mathematical Tables}.
\newblock Number~55 in National Bureau of Standards Applied Mathematics Series. U.S. Government Printing Office, Washington, D.C., 1964.

\bibitem[Bochem et~al.(2025)Bochem, Jung, Prasad, Conti, and Benini]{10992712}
Severin Bochem, Victor~J.B. Jung, Arpan~Suravi Prasad, Francesco Conti, and Luca Benini.
\newblock Distributed inference with minimal off-chip traffic for transformers on low-power {MCUs}.
\newblock In \emph{2025 Design, Automation \& Test in Europe Conference (DATE)}, pp.\  1--7, 2025.
\newblock \doi{10.23919/DATE64628.2025.10992712}.

\bibitem[Cao et~al.(2025)Cao, Peng, Gao, Hao, Li, and Mou]{cao2025object}
Jinfeng Cao, Bo~Peng, Mingzhong Gao, Haichun Hao, Xinfang Li, and Hongwei Mou.
\newblock Object detection based on cnn and vision-transformer: A survey.
\newblock \emph{IET Computer Vision}, 19\penalty0 (1):\penalty0 e70028, 2025.

\bibitem[Carion et~al.(2020)Carion, Massa, Synnaeve, Usunier, Kirillov, and Zagoruyko]{carion2020end}
Nicolas Carion, Francisco Massa, Gabriel Synnaeve, Nicolas Usunier, Alexander Kirillov, and Sergey Zagoruyko.
\newblock End-to-end object detection with transformers.
\newblock In \emph{European Conference on Computer Vision}, pp.\  213--229. Springer, 2020.

\bibitem[Chen et~al.(2024)Chen, Li, and Aly]{chen2024vita}
Chunyun Chen, Lantian Li, and Mohamed M~Sabry Aly.
\newblock Vita: A highly efficient dataflow and architecture for vision transformers.
\newblock In \emph{2024 Design, Automation \& Test in Europe Conference \& Exhibition (DATE)}, pp.\  1--6. IEEE, 2024.

\bibitem[Chen et~al.(2026)Chen, Lu, Zhu, Sun, and Liu]{chen2026stronger}
Mingzhi Chen, Taiming Lu, Jiachen Zhu, Mingjie Sun, and Zhuang Liu.
\newblock Stronger normalization-free transformers.
\newblock In \emph{Proceedings of the IEEE/CVF Conference on Computer Vision and Pattern Recognition (CVPR)}, 2026.

\bibitem[Chen et~al.(2017)Chen, Krishna, Emer, and Sze]{7738524}
Yu-Hsin Chen, Tushar Krishna, Joel~S. Emer, and Vivienne Sze.
\newblock Eyeriss: An energy-efficient reconfigurable accelerator for deep convolutional neural networks.
\newblock \emph{IEEE Journal of Solid-State Circuits}, 52\penalty0 (1):\penalty0 127--138, 2017.
\newblock \doi{10.1109/JSSC.2016.2616357}.

\bibitem[Chen et~al.(2025)Chen, Hao, Cao, Zhang, Shen, Li, and Zhang]{chen2025hardware}
Ziyang Chen, Ming Hao, Xinye Cao, Jingwei Zhang, Chaoyao Shen, Guoqing Li, and Meng Zhang.
\newblock Hardware-friendly and efficient vision transformer for deployment on low-power embedded device.
\newblock \emph{Journal of Low Power Electronics and Applications}, 16\penalty0 (1):\penalty0 1, 2025.

\bibitem[Dai et~al.(2026)Dai, Zou, Yang, Lin, Lou, Luo, Cai, Yin, Luk, and Wang]{LORA}
Yuan Dai, Guibin Zou, Yuanda Yang, Huan Lin, Jiahang Lou, Yiwen Luo, Xinyu Cai, Wenbo Yin, Wai-Shing Luk, and Lingli Wang.
\newblock {LoRA}: Towards improved applicability of reconfigurable architecture for versatile nonlinear functions.
\newblock In \emph{2026 ACM/IEEE 53rd Annual International Symposium on Computer Architecture (ISCA)}, pp.\  825--841, 2026.
\newblock \doi{10.1109/ISCA66397.2026.00068}.

\bibitem[De~Vries et~al.(2025)De~Vries, Keemink, and van Gerven]{de2025kozax}
Sigur De~Vries, Sander~Wessel Keemink, and Marcel Antonius~Johannes van Gerven.
\newblock Kozax: flexible and scalable genetic programming in jax.
\newblock In \emph{Proceedings of the Genetic and Evolutionary Computation Conference Companion}, pp.\  603--606, 2025.

\bibitem[Deb et~al.(2002)Deb, Pratap, Agarwal, and Meyarivan]{deb2002fast}
Kalyanmoy Deb, Amrit Pratap, Sameer Agarwal, and TAMT Meyarivan.
\newblock A fast and elitist multiobjective genetic algorithm: Nsga-ii.
\newblock \emph{IEEE Transactions on Evolutionary Computation}, 6\penalty0 (2):\penalty0 182--197, 2002.

\bibitem[Deng et~al.(2009)Deng, Dong, Socher, Li, Li, and Fei-Fei]{deng2009imagenet}
Jia Deng, Wei Dong, Richard Socher, Li-Jia Li, Kai Li, and Li~Fei-Fei.
\newblock Imagenet: A large-scale hierarchical image database.
\newblock In \emph{2009 IEEE Conference on Computer Vision and Pattern Recognition}, pp.\  248--255. Ieee, 2009.

\bibitem[Dong et~al.(2024)Dong, Zhuang, Yang, Ji, Li, Xu, Huang, Hu, Jones, Shi, et~al.]{dong2024eq}
Peiyan Dong, Jinming Zhuang, Zhuoping Yang, Shixin Ji, Yanyu Li, Dongkuan Xu, Heng Huang, Jingtong Hu, Alex~K Jones, Yiyu Shi, et~al.
\newblock Eq-vit: Algorithm-hardware co-design for end-to-end acceleration of real-time vision transformer inference on versal acap architecture.
\newblock \emph{IEEE Transactions on Computer-Aided Design of Integrated Circuits and Systems}, 43\penalty0 (11):\penalty0 3949--3960, 2024.

\bibitem[Dosovitskiy et~al.(2020)Dosovitskiy, Beyer, Kolesnikov, Weissenborn, Zhai, Unterthiner, Dehghani, Minderer, Heigold, Gelly, et~al.]{dosovitskiy2020image}
Alexey Dosovitskiy, Lucas Beyer, Alexander Kolesnikov, Dirk Weissenborn, Xiaohua Zhai, Thomas Unterthiner, Mostafa Dehghani, Matthias Minderer, Georg Heigold, Sylvain Gelly, et~al.
\newblock An image is worth 16x16 words: Transformers for image recognition at scale.
\newblock \emph{arXiv preprint arXiv:2010.11929}, 2020.

\bibitem[Dumoulin et~al.(2024)Dumoulin, Houshmand, Jain, and Verhelst]{10558587}
Joren Dumoulin, Pouya Houshmand, Vikram Jain, and Marian Verhelst.
\newblock Enabling efficient hardware acceleration of hybrid vision transformer (vit) networks at the edge.
\newblock In \emph{2024 IEEE International Symposium on Circuits and Systems (ISCAS)}, pp.\  1--5, 2024.
\newblock \doi{10.1109/ISCAS58744.2024.10558587}.

\bibitem[{Efabless Corporation}()]{caravel}
{Efabless Corporation}.
\newblock Caravel: A template {SoC} for the {Open MPW} program.
\newblock \url{https://github.com/efabless/caravel}.
\newblock Accessed: 2026-11-13.

\bibitem[Hendrycks et~al.(2021{\natexlab{a}})Hendrycks, Basart, Mu, Kadavath, Wang, Dorundo, Desai, Zhu, Parajuli, Guo, Song, Steinhardt, and Gilmer]{hendrycks2021many}
Dan Hendrycks, Steven Basart, Norman Mu, Saurav Kadavath, Frank Wang, Evan Dorundo, Rahul Desai, Tyler Zhu, Samyak Parajuli, Mike Guo, Dawn Song, Jacob Steinhardt, and Justin Gilmer.
\newblock The many faces of robustness: A critical analysis of out-of-distribution generalization.
\newblock In \emph{Proceedings of the IEEE/CVF International Conference on Computer Vision (ICCV)}, pp.\  8340--8349, 2021{\natexlab{a}}.

\bibitem[Hendrycks et~al.(2021{\natexlab{b}})Hendrycks, Zhao, Basart, Steinhardt, and Song]{hendrycks2021natural}
Dan Hendrycks, Kevin Zhao, Steven Basart, Jacob Steinhardt, and Dawn Song.
\newblock Natural adversarial examples.
\newblock In \emph{Proceedings of the IEEE/CVF Conference on Computer Vision and Pattern Recognition (CVPR)}, pp.\  15262--15271, 2021{\natexlab{b}}.

\bibitem[Higham(2002)]{10.5555/579525}
Nicholas~J. Higham.
\newblock \emph{Accuracy and Stability of Numerical Algorithms}.
\newblock Society for Industrial and Applied Mathematics, 2nd edition, 2002.
\newblock ISBN 0898715210.

\bibitem[Hinton et~al.(2015)Hinton, Vinyals, and Dean]{hinton2015distilling}
Geoffrey Hinton, Oriol Vinyals, and Jeff Dean.
\newblock Distilling the knowledge in a neural network.
\newblock \emph{arXiv preprint arXiv:1503.02531}, 2015.

\bibitem[Kim et~al.(2025)Kim, Lee, Kim, Park, and Lee]{kim2025hardware}
Raehyeong Kim, Dayoung Lee, Jinyeol Kim, Joungmin Park, and Seung~Eun Lee.
\newblock Hardware accelerator for approximation-based softmax and layer normalization in transformers.
\newblock \emph{Electronics}, 14\penalty0 (12):\penalty0 2337, 2025.

\bibitem[Kim et~al.(2024)Kim, Kim, Jo, Kim, Hong, and Yoo]{10454330}
Sangyeob Kim, Sangjin Kim, Wooyoung Jo, Soyeon Kim, Seongyon Hong, and Hoi-Jun Yoo.
\newblock C-transformer: A {2.6-18.1$\mu$J/Token} homogeneous {DNN}-transformer/spiking-transformer processor with big-little network and implicit weight generation for large language models.
\newblock In \emph{2024 IEEE International Solid-State Circuits Conference (ISSCC)}, volume~67, pp.\  368--370, 2024.
\newblock \doi{10.1109/ISSCC49657.2024.10454330}.

\bibitem[Koza(1994)]{koza1994genetic}
John~R Koza.
\newblock Genetic programming as a means for programming computers by natural selection.
\newblock \emph{Statistics and Computing}, 4\penalty0 (2):\penalty0 87--112, 1994.

\bibitem[Lee et~al.(2024)Lee, Kwon, Park, Yu, Park, and Song]{lee2024q}
Jemin Lee, Yongin Kwon, Sihyeong Park, Misun Yu, Jeman Park, and Hwanjun Song.
\newblock Q-hyvit: Post-training quantization of hybrid vision transformers with bridge block reconstruction for iot systems.
\newblock \emph{IEEE Internet of Things Journal}, 11\penalty0 (22):\penalty0 36384--36396, 2024.

\bibitem[Lee et~al.(2026)Lee, Das, Han, and Chandrakasan]{11343916}
Kyungmi Lee, Gaurab Das, Donghyeon Han, and Anantha~P. Chandrakasan.
\newblock Securing {DNN} acceleration from off-chip memory vulnerabilities with low-overhead authenticated encryption.
\newblock \emph{IEEE Transactions on Very Large Scale Integration (VLSI) Systems}, 34\penalty0 (3):\penalty0 953--966, 2026.
\newblock \doi{10.1109/TVLSI.2025.3650411}.

\bibitem[Liu et~al.(2023)Liu, Peng, Zheng, Yang, Hu, and Yuan]{liu2023efficientvit}
Xinyu Liu, Houwen Peng, Ningxin Zheng, Yuqing Yang, Han Hu, and Yixuan Yuan.
\newblock Efficientvit: Memory efficient vision transformer with cascaded group attention.
\newblock In \emph{2023 IEEE/CVF Conference on Computer Vision and Pattern Recognition (CVPR)}, pp.\  14420--14430. Ieee, 2023.

\bibitem[Liu et~al.(2021)Liu, Lin, Cao, Hu, Wei, Zhang, Lin, and Guo]{liu2021swin}
Ze~Liu, Yutong Lin, Yue Cao, Han Hu, Yixuan Wei, Zheng Zhang, Stephen Lin, and Baining Guo.
\newblock Swin transformer: Hierarchical vision transformer using shifted windows.
\newblock In \emph{Proceedings of the IEEE/CVF International Conference on Computer Vision}, pp.\  10012--10022, 2021.

\bibitem[Liu et~al.(2022)Liu, Mao, Wu, Feichtenhofer, Darrell, and Xie]{liu2022convnet}
Zhuang Liu, Hanzi Mao, Chao-Yuan Wu, Christoph Feichtenhofer, Trevor Darrell, and Saining Xie.
\newblock A {ConvNet} for the 2020s.
\newblock In \emph{Proceedings of the IEEE/CVF Conference on Computer Vision and Pattern Recognition}, pp.\  11976--11986, 2022.

\bibitem[Lokhande et~al.(2025)Lokhande, Raut, and Vishvakarma]{FlexPE}
Mukul Lokhande, Gopal Raut, and Santosh~Kumar Vishvakarma.
\newblock {Flex-PE}: Flexible and {SIMD} multiprecision processing element for {AI} workloads.
\newblock \emph{IEEE Transactions on Very Large Scale Integration (VLSI) Systems}, 33\penalty0 (6):\penalty0 1610--1623, 2025.
\newblock \doi{10.1109/TVLSI.2025.3553069}.

\bibitem[Marino et~al.(2023)Marino, Zhang, and Prasanna]{marino2023me}
Kyle Marino, Pengmiao Zhang, and Viktor~K Prasanna.
\newblock Me-vit: A single-load memory-efficient fpga accelerator for vision transformers.
\newblock In \emph{2023 IEEE 30th International Conference on High Performance Computing, Data, and Analytics (HiPC)}, pp.\  213--223. IEEE, 2023.

\bibitem[Moon et~al.(2025)Moon, Li, Chen, Knag, Krishnamurthy, and Seok]{10904793}
Seunghyun Moon, Mao Li, Gregory~K. Chen, Phil~C. Knag, Ram~Kumar Krishnamurthy, and Mingoo Seok.
\newblock {T-REX}: A {68-to-567$\mu$s/Token 0.41-to-3.95$\mu$J/Token} transformer accelerator with reduced external memory access and enhanced hardware utilization in 16nm {FinFET}.
\newblock In \emph{2025 IEEE International Solid-State Circuits Conference (ISSCC)}, volume~68, pp.\  406--408, 2025.
\newblock \doi{10.1109/ISSCC49661.2025.10904793}.

\bibitem[Muller et~al.(2018)Muller, Brunie, De~Dinechin, Jeannerod, Joldes, Lef{\`e}vre, Melquiond, Revol, and Torres]{muller2018handbook}
Jean-Michel Muller, Nicolas Brunie, Florent De~Dinechin, Claude-Pierre Jeannerod, Mioara Joldes, Vincent Lef{\`e}vre, Guillaume Melquiond, Nathalie Revol, and Serge Torres.
\newblock \emph{Handbook of Floating-Point Arithmetic}, volume~1.
\newblock Springer, 2018.

\bibitem[Nag et~al.(2023)Nag, Datta, Kundu, Chandrachoodan, and Beerel]{10181988}
Shashank Nag, Gourav Datta, Souvik Kundu, Nitin Chandrachoodan, and Peter~A. Beerel.
\newblock {ViTA}: A vision transformer inference accelerator for edge applications.
\newblock In \emph{2023 IEEE International Symposium on Circuits and Systems (ISCAS)}, pp.\  1--5, 2023.
\newblock \doi{10.1109/ISCAS46773.2023.10181988}.

\bibitem[Ou et~al.(2023)Ou, Wu, Wang, Chen, and Yang]{COMPACT}
Wenhui Ou, Zhuoyu Wu, Zheng Wang, Chao Chen, and Yongkui Yang.
\newblock {COMPACT}: Co-processor for multi-mode precision-adjustable non-linear activation functions.
\newblock In \emph{2023 Design, Automation \& Test in Europe Conference \& Exhibition ({DATE})}, pp.\  1--6, 2023.
\newblock \doi{10.23919/DATE56975.2023.10137019}.

\bibitem[Pan(1966)]{pan1966methods}
V~Ya Pan.
\newblock Methods of computing values of polynomials.
\newblock \emph{Russian Mathematical Surveys}, 21\penalty0 (1):\penalty0 105--136, 1966.

\bibitem[Qiu et~al.(2026)Qiu, Li, Wu, Jia, Ye, and Ma]{11130436}
Yikan Qiu, Guoxiang Li, Meng Wu, Yifan Jia, Le~Ye, and Yufei Ma.
\newblock Quartet: A digital compute-in-memory versatile {AI} accelerator with heterogeneous tensor engines and off-chip-less dataflow.
\newblock \emph{IEEE Transactions on Circuits and Systems I: Regular Papers}, 73\penalty0 (1):\penalty0 370--383, 2026.
\newblock \doi{10.1109/TCSI.2025.3598287}.

\bibitem[Raghu et~al.(2021)Raghu, Unterthiner, Kornblith, Zhang, and Dosovitskiy]{raghu2021vision}
Maithra Raghu, Thomas Unterthiner, Simon Kornblith, Chiyuan Zhang, and Alexey Dosovitskiy.
\newblock Do vision transformers see like convolutional neural networks?
\newblock \emph{Advances in Neural Information Processing Systems}, 34:\penalty0 12116--12128, 2021.

\bibitem[Recht et~al.(2019)Recht, Roelofs, Schmidt, and Shankar]{recht2019imagenet}
Benjamin Recht, Rebecca Roelofs, Ludwig Schmidt, and Vaishaal Shankar.
\newblock Do {ImageNet} classifiers generalize to {ImageNet}?
\newblock In \emph{International Conference on Machine Learning (ICML)}, pp.\  5389--5400, 2019.

\bibitem[Sadeghi et~al.(2024)Sadeghi, Fayyazi, Azizi, and Pedram]{sadeghi2024peano}
Mohammad~Erfan Sadeghi, Arash Fayyazi, Seyedarmin Azizi, and Massoud Pedram.
\newblock Peano-vit: Power-efficient approximations of non-linearities in vision transformers.
\newblock In \emph{Proceedings of the 29th ACM/IEEE International Symposium on Low Power Electronics and Design}, pp.\  1--6, 2024.

\bibitem[Satopaa et~al.(2011)Satopaa, Albrecht, Irwin, and Raghavan]{satopaa2011finding}
Ville Satopaa, Jeannie Albrecht, David Irwin, and Barath Raghavan.
\newblock Finding a" kneedle" in a haystack: Detecting knee points in system behavior.
\newblock In \emph{2011 31st International Conference on Distributed Computing Systems Workshops}, pp.\  166--171. IEEE, 2011.

\bibitem[Sun et~al.(2025)Sun, Ma, Liu, Li, Li, Wang, Liu, and Liu]{sun2025integer}
Tianyi Sun, Tuo Ma, Jiali Liu, Zhiwei Li, Qingjiang Li, Yinan Wang, Haijun Liu, and Sen Liu.
\newblock Integer quantization of nonlinear operations towards hardware-friendly vits.
\newblock In \emph{2025 32nd IEEE International Conference on Electronics, Circuits and Systems (ICECS)}, pp.\  1--4. IEEE, 2025.

\bibitem[Tabani et~al.(2021)Tabani, Balasubramaniam, Marzban, Arani, and Zonooz]{tabani2021improving}
Hamid Tabani, Ajay Balasubramaniam, Shabbir Marzban, Elahe Arani, and Bahram Zonooz.
\newblock Improving the efficiency of transformers for resource-constrained devices.
\newblock In \emph{2021 24th Euromicro Conference on Digital System Design (DSD)}, pp.\  449--456. IEEE, 2021.

\bibitem[Tang(1989)]{10.1145/63522.214389}
Ping-Tak~Peter Tang.
\newblock Table-driven implementation of the exponential function in {IEEE} floating-point arithmetic.
\newblock \emph{ACM Trans. Math. Softw.}, 15\penalty0 (2):\penalty0 144--157, June 1989.
\newblock ISSN 0098-3500.
\newblock \doi{10.1145/63522.214389}.
\newblock URL \url{https://doi.org/10.1145/63522.214389}.

\bibitem[Thisanke et~al.(2023)Thisanke, Deshan, Chamith, Seneviratne, Vidanaarachchi, and Herath]{thisanke2023semantic}
Hans Thisanke, Chamli Deshan, Kavindu Chamith, Sachith Seneviratne, Rajith Vidanaarachchi, and Damayanthi Herath.
\newblock Semantic segmentation using vision transformers: A survey.
\newblock \emph{Engineering Applications of Artificial Intelligence}, 126:\penalty0 106669, 2023.

\bibitem[Touvron et~al.(2021)Touvron, Cord, Douze, Massa, Sablayrolles, and J{\'e}gou]{touvron2021training}
Hugo Touvron, Matthieu Cord, Matthijs Douze, Francisco Massa, Alexandre Sablayrolles, and Herv{\'e} J{\'e}gou.
\newblock Training data-efficient image transformers \& distillation through attention.
\newblock In \emph{International Conference on Machine Learning}, pp.\  10347--10357. PMLR, 2021.

\bibitem[Wang et~al.(2019)Wang, Ge, Lipton, and Xing]{wang2019learning}
Haohan Wang, Songwei Ge, Zachary Lipton, and Eric~P. Xing.
\newblock Learning robust global representations by penalizing local predictive power.
\newblock In \emph{Advances in Neural Information Processing Systems (NeurIPS)}, 2019.

\bibitem[Wang et~al.(2023)Wang, Zhou, Sun, Sun, and Liu]{wang2023sole}
Wenxun Wang, Shuchang Zhou, Wenyu Sun, Peiqin Sun, and Yongpan Liu.
\newblock Sole: Hardware-software co-design of softmax and layernorm for efficient transformer inference.
\newblock In \emph{2023 IEEE/ACM International Conference on Computer Aided Design (ICCAD)}, pp.\  1--9. IEEE, 2023.

\bibitem[Wightman(2019)]{rw2019timm}
Ross Wightman.
\newblock Pytorch image models.
\newblock \url{https://github.com/rwightman/pytorch-image-models}, 2019.

\bibitem[Xiao et~al.(2025)Xiao, Cheng, and Zhao]{xiao2025refining}
Can Xiao, Jianyi Cheng, and Yiren Zhao.
\newblock Refining datapath for microscaling vits.
\newblock In \emph{2025 35th International Conference on Field-Programmable Logic and Applications (FPL)}, pp.\  263--272. IEEE, 2025.

\bibitem[Xu et~al.(2025)Xu, Banerjee, and Gupta]{xu2025hardware}
Bin Xu, Ayan Banerjee, and Sandeep Gupta.
\newblock Hardware acceleration for neural networks: A comprehensive survey.
\newblock \emph{arXiv preprint arXiv:2512.23914}, 2025.

\bibitem[Yu et~al.(2022)Yu, Park, Park, Kim, Lee, Lee, and Choi]{yu2022nn}
Joonsang Yu, Junki Park, Seongmin Park, Minsoo Kim, Sihwa Lee, Dong~Hyun Lee, and Jungwook Choi.
\newblock Nn-lut: Neural approximation of non-linear operations for efficient transformer inference.
\newblock In \emph{Proceedings of the 59th ACM/IEEE Design Automation Conference}, pp.\  577--582, 2022.

\bibitem[Zhao et~al.(2025)Zhao, Li, Liu, Lu, Wang, Yang, Jiang, and Guan]{zhao2025quark}
Zhixiong Zhao, Haomin Li, Fangxin Liu, Yuncheng Lu, Zongwu Wang, Tao Yang, Li~Jiang, and Haibing Guan.
\newblock Quark: Quantization-enabled circuit sharing for transformer acceleration by exploiting common patterns in nonlinear operations.
\newblock In \emph{2025 IEEE/ACM International Conference On Computer Aided Design (ICCAD)}, pp.\  1--9. IEEE, 2025.

\bibitem[Zhu et~al.(2025)Zhu, Chen, He, LeCun, and Liu]{zhu2025transformers}
Jiachen Zhu, Xinlei Chen, Kaiming He, Yann LeCun, and Zhuang Liu.
\newblock Transformers without normalization.
\newblock In \emph{Proceedings of the Computer Vision and Pattern Recognition Conference}, pp.\  14901--14911, 2025.

\end{thebibliography}
\bibliographystyle{iclr2027_conference}

\appendix

\section{Normalization mappings}
\label{A:mappings}
\begin{figure}[htbp]
    \centering
    \includegraphics[width=\textwidth]{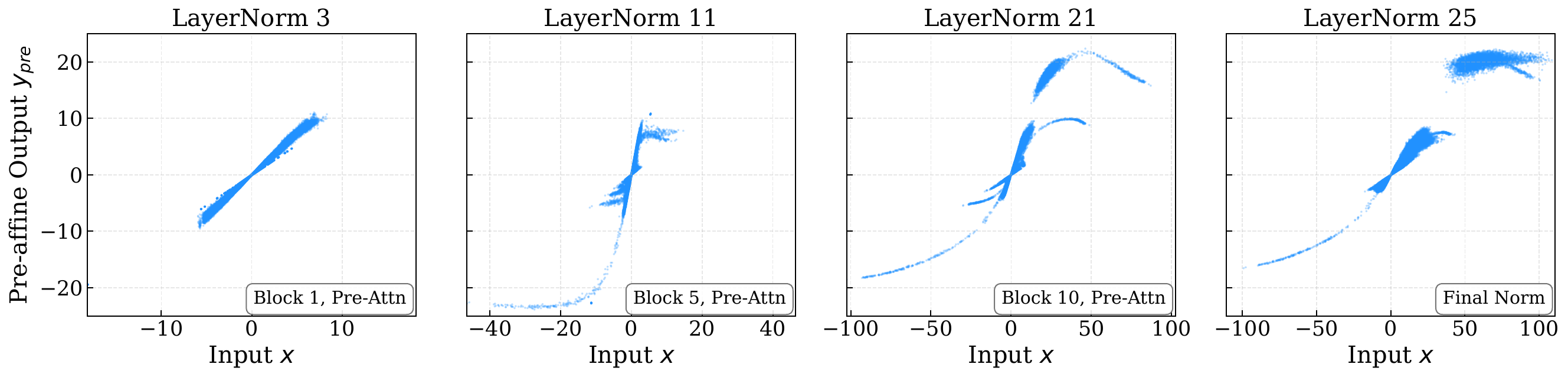}
    \caption{\textbf{Evolution of LayerNorm mappings across network depth in a pre-trained ViT-B.} Scatter plots visualize the pre-affine output $y_{pre}$ against the input $x$ across four representative layers. Early normalization blocks (e.g., LayerNorm 3 and 11) exhibit largely linear transformations, whereas deeper blocks (e.g., LayerNorm 21 and 25) display highly non-linear, S-shaped characteristics. This distinct structural variation across the network motivates our layer-specific symbolic discovery approach to accurately capture the diverse behaviours across the network.}
    \label{fig:mappings}
\end{figure}

Figure~\ref{fig:mappings} visualizes the underlying distribution of the pre-affine normalization mappings across four representative layers. Generated using 5 million sampled data points, these plots clearly illustrate the network's transition from near-linear behaviour in early blocks to highly non-linear, S-shaped curves in deeper blocks.

\section{Experimental settings}
\label{A:exp_settings}
All re-alignment experiments are conducted on the ImageNet-1K dataset using pre-trained ViT-B/16 (\texttt{vit\_base\_patch16\_224}), ViT-L/16 (\texttt{vit\_large\_patch16\_224}), Swin-B (\texttt{swin\_base\_patch4\_window7\_224}), and ConvNeXt-B (\texttt{convnext\_base}) architectures. The following settings were held constant across all variants to ensure a fair comparison:

\begin{itemize}
    \item \textbf{Hardware and precision:} ViT-B experiments were performed on a single NVIDIA A100 (80GB) GPU. ViT-L experiments were trained on 2 GPUs via data-parallel training, using either NVIDIA A100 (80GB) or H100 (80GB) GPUs. Swin-B and ConvNeXt-B experiments each used a single GPU, an NVIDIA A100 (80GB) for Swin-B and an NVIDIA H100 (80GB) for ConvNeXt-B. We utilized Automatic Mixed Precision (AMP) throughout to optimize memory usage and training speed. AMP used float16, except for the GP variants on ConvNeXt-B and one GP seed on Swin-B, which overflowed in float16 and used bfloat16 instead.
    \item \textbf{Optimization:} We used the AdamW optimizer with default momentum parameters $(\beta_1, \beta_2) = (0.9, 0.999)$. The global batch size was set to 512. Each variant was fine-tuned for exactly 20 epochs without a warmup phase, using the learning rate schedule listed for that variant in Tables~\ref{tab:appendix_hparams_vitb}--\ref{tab:appendix_hparams_convnextb}.
    \item \textbf{Data augmentation:} We adopted the standard augmentation strategy used in~\citet{touvron2021training}. Specifically, we applied RandAugment (\texttt{rand-m9-mstd0.5-inc1}), Color Jitter (0.4), and Random Erasing (0.25). Consistent with a short fine-tuning regime, we did not employ Mixup, Cutmix, or Label Smoothing.
    \item \textbf{Evaluation:} Validation performance was measured using a Center Crop with a crop ratio of 0.875 (0.9 for Swin-B).
    \item \textbf{Distillation configuration:} For all distillation variants (GP-D, DyT-D, and Derf-D), we utilized logit-based distillation with a temperature $\tau = 4.0$ and a loss balancing coefficient $\lambda = 0.5$.
    \item \textbf{Compute time and total resources:} Depending on the specific architectural variant and cluster node allocation, a standard 20-epoch fine-tuning run takes approximately 4--9 hours for ViT-B, 6--16 hours for ViT-L, 18--28 hours for Swin-B, and 6--10 hours for ConvNeXt-B. The GP search (Section~\ref{symbolic_search}) runs on a single NVIDIA A100 (80GB) GPU and takes approximately 1.6 hours per seed for ViT-B, 3.7 hours for ViT-L, 2.8 hours for Swin-B, and 2.1 hours for ConvNeXt-B (3--5 minutes per normalization layer). Accounting for the evaluation of all variants, hyperparameter sweeps, preliminary trials, and final logging across all four architectures, we estimate the total compute required for this research project to be approximately 4{,}800 GPU hours.
\end{itemize}

\section{Hyperparameters}
\label{A:hyperparams}
To ensure a fair comparison, we performed an independent grid search for each variant to identify the optimal learning rate, weight decay, and stochastic depth rate on ViT-B and ViT-L. For Swin-B and ConvNeXt-B, we searched only the learning rate and kept the weight decay and stochastic depth of the ViT-B runs. The final configurations are summarized in Tables~\ref{tab:appendix_hparams_vitb}--\ref{tab:appendix_hparams_convnextb}. As described in Section~\ref{model_realign}, the DyT-F and DyT-D variants were highly sensitive to hyperparameters at ViT-B and ViT-L. For these variants, independent learning rates were applied to the backbone, affine parameters, and the learnable $\alpha$ scalars. For Swin-B and ConvNeXt-B, the search scaled these three learning rates by a common factor, which gave a factor of $1$ for Swin-B and $3$ for ConvNeXt-B. In contrast, the GP variants were successfully optimized using a single global learning rate for all trainable parameters. The Derf variants use the hyperparameters of the corresponding DyT variant on each architecture, with all learning rates scaled by a common factor. We searched this factor over $1/3$, $1$, and $3$, adding one further step ($9$ or $1/9$) where the best value lay at the edge of this range. This gave a factor of $3$ for Derf-F at ViT-B, $1/3$ for Derf-A at ViT-L, and $1$ in all other cases. As for DyT-A, $\alpha$ and $s$ are not trained in Derf-A and keep their initial values of $0.5$ and $0$.

\begin{table}[htbp]
\centering
\caption{\textbf{Optimal hyperparameters for ImageNet-1K re-alignment (ViT-B).} Parameters were determined via independent grid searches for each variant to ensure a fair performance comparison. For variants where the backbone is frozen (-A), the backbone learning rate is denoted with a hyphen (-). The $LR_{\text{affine}}$ row corresponds to the learning rate for the affine transformation parameters (weights and biases), which were inherited from the original pre-trained normalization layers. For DyT variants, $LR_{\alpha}$ denotes the specific learning rate for the learnable scaling parameter. For DyT-A, $\alpha$ is not trained and keeps its initial value of $0.5$. $^\dagger$Backbone weight decay; the DyT affine parameters used $0.0$.}
\label{tab:appendix_hparams_vitb}
\begin{adjustbox}{width=\textwidth}
\begin{tabular}{lcccccc}
    \toprule
    \textbf{Parameter} & \textbf{GP-A} & \textbf{DyT-A} & \textbf{GP-F} & \textbf{DyT-F} & \textbf{GP-D} & \textbf{DyT-D} \\
    \midrule
    $LR_{\text{backbone}}$ & - & - & $1 \times 10^{-5}$ & $2 \times 10^{-5}$ & $1 \times 10^{-5}$ & $3 \times 10^{-5}$ \\
    $LR_{\text{affine}}$   & $1 \times 10^{-3}$ & $8 \times 10^{-3}$ & $1 \times 10^{-5}$ & $1 \times 10^{-4}$ & $1 \times 10^{-5}$ & $1 \times 10^{-4}$ \\
    $LR_{\alpha}$          & - & - & - & $5 \times 10^{-5}$ & - & $5 \times 10^{-5}$ \\
    LR Schedule            & Constant & Cosine & Cosine & Cosine & Cosine & Cosine \\
    Weight Decay           & $0.0$ & $0.0$ & $0.05$ & $0.02^\dagger$ & $0.05$ & $0.02^\dagger$ \\
    Stoch. Depth           & $0.0$ & $0.0$ & $0.2$ & $0.0$ & $0.2$ & $0.0$ \\
    \bottomrule
\end{tabular}
\end{adjustbox}
\end{table}

\begin{table}[htbp]
\centering
\caption{\textbf{Optimal hyperparameters for ImageNet-1K re-alignment (ViT-L).} Parameter definitions and notation follow Table~\ref{tab:appendix_hparams_vitb}. $^\dagger$Backbone weight decay; the DyT affine parameters used $0.05$.}
\label{tab:appendix_hparams_vitl}
\begin{adjustbox}{width=\textwidth}
\begin{tabular}{lcccccc}
    \toprule
    \textbf{Parameter} & \textbf{GP-A} & \textbf{DyT-A} & \textbf{GP-F} & \textbf{DyT-F} & \textbf{GP-D} & \textbf{DyT-D} \\
    \midrule
    $LR_{\text{backbone}}$ & - & - & $1 \times 10^{-5}$ & $2 \times 10^{-5}$ & $1 \times 10^{-5}$ & $3 \times 10^{-5}$ \\
    $LR_{\text{affine}}$   & $2 \times 10^{-3}$ & $8 \times 10^{-3}$ & $1 \times 10^{-5}$ & $1 \times 10^{-4}$ & $1 \times 10^{-5}$ & $1 \times 10^{-4}$ \\
    $LR_{\alpha}$          & - & - & - & $5 \times 10^{-5}$ & - & $5 \times 10^{-5}$ \\
    LR Schedule            & Cosine & Cosine & Cosine & Cosine & Cosine & Cosine \\
    Weight Decay           & $0.0$ & $0.0$ & $0.05$ & $0.02^\dagger$ & $0.05$ & $0.02^\dagger$ \\
    Stoch. Depth           & $0.0$ & $0.0$ & $0.2$ & $0.2$ & $0.2$ & $0.2$ \\
    \bottomrule
\end{tabular}
\end{adjustbox}
\end{table}

\begin{table}[htbp]
\centering
\caption{\textbf{Optimal hyperparameters for ImageNet-1K re-alignment (Swin-B).} Parameter definitions and notation follow Table~\ref{tab:appendix_hparams_vitb}. $^\dagger$Backbone weight decay; the DyT affine parameters used $0.0$.}
\label{tab:appendix_hparams_swinb}
\begin{adjustbox}{width=\textwidth}
\begin{tabular}{lcccccc}
    \toprule
    \textbf{Parameter} & \textbf{GP-A} & \textbf{DyT-A} & \textbf{GP-F} & \textbf{DyT-F} & \textbf{GP-D} & \textbf{DyT-D} \\
    \midrule
    $LR_{\text{backbone}}$ & - & - & $3 \times 10^{-5}$ & $2 \times 10^{-5}$ & $3 \times 10^{-5}$ & $3 \times 10^{-5}$ \\
    $LR_{\text{affine}}$   & $3 \times 10^{-3}$ & $3 \times 10^{-3}$ & $3 \times 10^{-5}$ & $1 \times 10^{-4}$ & $3 \times 10^{-5}$ & $1 \times 10^{-4}$ \\
    $LR_{\alpha}$          & - & - & - & $5 \times 10^{-5}$ & - & $5 \times 10^{-5}$ \\
    LR Schedule            & Cosine & Cosine & Cosine & Cosine & Cosine & Cosine \\
    Weight Decay           & $0.0$ & $0.0$ & $0.05$ & $0.02^\dagger$ & $0.05$ & $0.02^\dagger$ \\
    Stoch. Depth           & $0.0$ & $0.0$ & $0.2$ & $0.0$ & $0.2$ & $0.0$ \\
    \bottomrule
\end{tabular}
\end{adjustbox}
\end{table}

\begin{table}[htbp]
\centering
\caption{\textbf{Optimal hyperparameters for ImageNet-1K re-alignment (ConvNeXt-B).} Parameter definitions and notation follow Table~\ref{tab:appendix_hparams_vitb}. $^\dagger$Backbone weight decay; the DyT affine parameters used $0.0$.}
\label{tab:appendix_hparams_convnextb}
\begin{adjustbox}{width=\textwidth}
\begin{tabular}{lcccccc}
    \toprule
    \textbf{Parameter} & \textbf{GP-A} & \textbf{DyT-A} & \textbf{GP-F} & \textbf{DyT-F} & \textbf{GP-D} & \textbf{DyT-D} \\
    \midrule
    $LR_{\text{backbone}}$ & - & - & $3 \times 10^{-5}$ & $6 \times 10^{-5}$ & $3 \times 10^{-5}$ & $9 \times 10^{-5}$ \\
    $LR_{\text{affine}}$   & $3 \times 10^{-3}$ & $3 \times 10^{-3}$ & $3 \times 10^{-5}$ & $3 \times 10^{-4}$ & $3 \times 10^{-5}$ & $3 \times 10^{-4}$ \\
    $LR_{\alpha}$          & - & - & - & $1.5 \times 10^{-4}$ & - & $1.5 \times 10^{-4}$ \\
    LR Schedule            & Cosine & Cosine & Cosine & Cosine & Cosine & Cosine \\
    Weight Decay           & $0.0$ & $0.0$ & $0.05$ & $0.02^\dagger$ & $0.05$ & $0.02^\dagger$ \\
    Stoch. Depth           & $0.0$ & $0.0$ & $0.2$ & $0.0$ & $0.2$ & $0.0$ \\
    \bottomrule
\end{tabular}
\end{adjustbox}
\end{table}

\newpage
\section{Complete LayerNorm mappings and discovered symbolic solutions}
\label{A:equations}

This section provides a comprehensive overview of the layer-specific symbolic solutions discovered by our GP framework, for all four architectures evaluated in this work.

\subsection{ViT-B}

\begin{figure}[htbp]
    \centering
    \includegraphics[width=\linewidth]{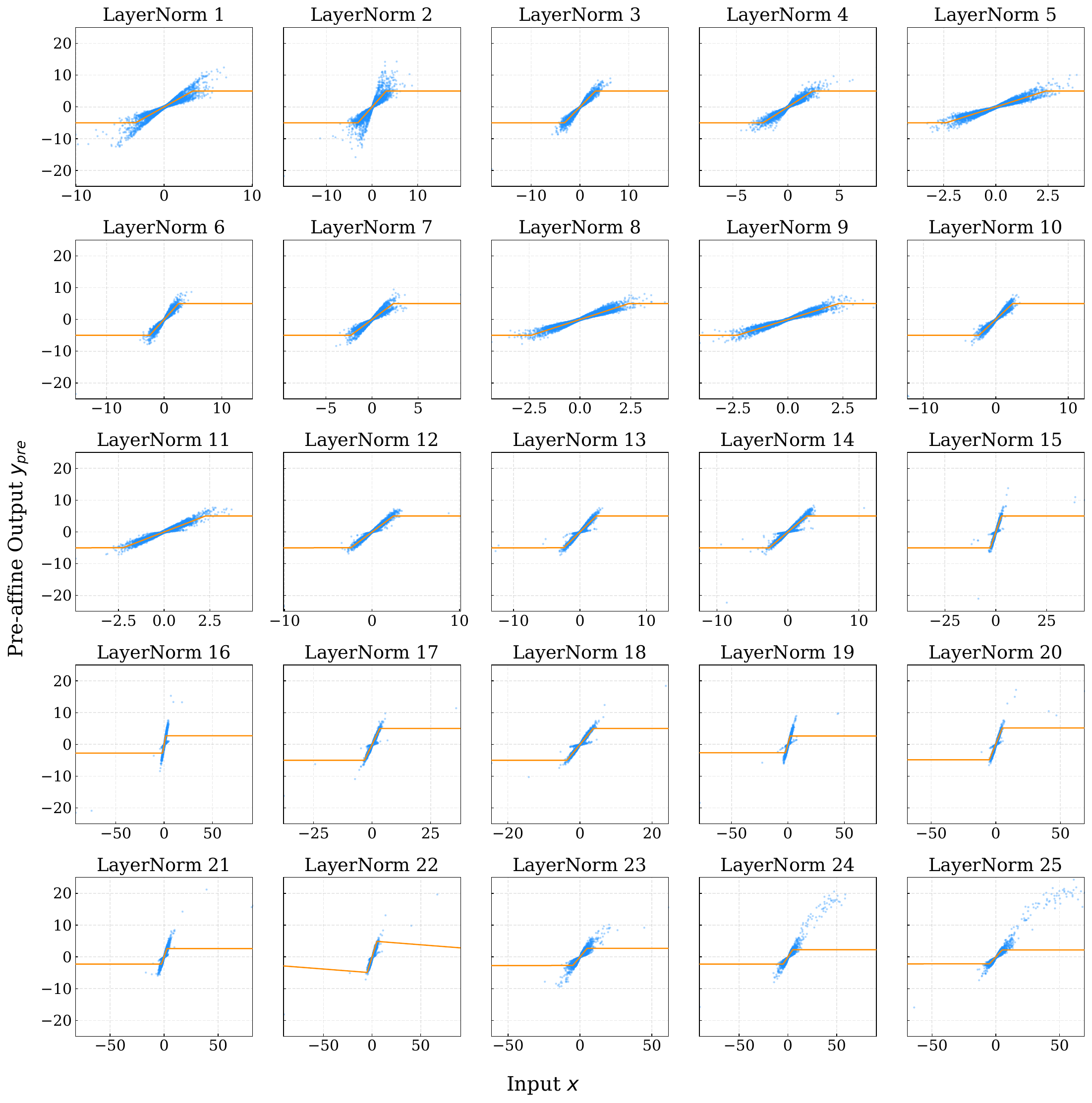}
    \caption{\textbf{Comprehensive functional alignment across all 25 normalization layers of ViT-B.} The evolved GP solutions (orange lines) are overlaid onto the target 50{,}000-point LayerNorm mappings (blue scatter) extracted from the pre-trained ViT-B architecture. The visualization demonstrates the framework's ability to seamlessly adapt to the structural transition from near-linear behaviour in early layers to highly non-linear, S-shaped characteristics in deeper blocks.}
    \label{fig:all25_mappings}
\end{figure}

Figure~\ref{fig:all25_mappings} visualizes the complete set of 25 normalization mappings extracted from the pre-trained ViT-B architecture, overlaid with the functional approximations discovered by the GP search. Correspondingly, Table~\ref{tab:gp_solutions_vitb} lists the exact mathematical formulations of these symbolic expressions. These solutions are derived from a single representative evolutionary search seed, where the specific expression for each layer was selected using the Kneedle algorithm to locate the optimal balance between functional fitness and computational cost on that layer's respective Pareto front.

\begin{table}[htbp]
\centering
\caption{\textbf{Optimal GP-evolved symbolic expressions per layer (ViT-B).} For each LayerNorm layer in the pre-trained ViT-B/16 architecture, we show the discovered symbolic solutions obtained from a single representative evolutionary seed (out of five independent runs). Solutions were selected by applying the Kneedle algorithm to locate the optimal balance of fitness and hardware cost on the Pareto front. \textbf{Train/Val Fitness} denotes the score achieved via the GP fitness function. \textbf{FLOPs} reports the per-token cost at hidden dimension $d$ under the unified $\exp$ primitive (Appendix~\ref{A:Hardware-centric}), where each $\tanh(x)$ counts as $23$ FLOPs, each $\sigma(x)$ as $22$, and each \texttt{clip} as zero (reducing to a comparison and a select instead of a floating-point operation).}
\label{tab:gp_solutions_vitb}
\resizebox{\textwidth}{!}{
\begin{tabular}{llccll}
\toprule
\# & Layer & Train Fitness ($\downarrow$) & Val Fitness ($\downarrow$) & Expression & FLOPs\\
\midrule
1  & \texttt{blocks.0.norm1}  & 0.2445 & 0.2362 & $\mathrm{clip}(1.54 x)$ & $1d$ \\
2  & \texttt{blocks.0.norm2}  & 0.3327 & 0.2732 & $\mathrm{clip}(1.62 x)$ & $1d$ \\
3  & \texttt{blocks.1.norm1}  & 0.1808 & 0.1739 & $\mathrm{clip}(1.58 x)$ & $1d$ \\
4  & \texttt{blocks.1.norm2}  & 0.1941 & 0.1912 & $\mathrm{clip}(2.08 x)$ & $1d$ \\
5  & \texttt{blocks.2.norm1}  & 0.2046 & 0.2029 & $\mathrm{clip}(2.1 x)$ & $1d$ \\
6  & \texttt{blocks.2.norm2}  & 0.2095 & 0.2020 & $\mathrm{clip}(2 x)$ & $1d$ \\
7  & \texttt{blocks.3.norm1}  & 0.2277 & 0.2248 & $\mathrm{clip}(2.17 x)$ & $1d$ \\
8  & \texttt{blocks.3.norm2}  & 0.2094 & 0.2093 & $\mathrm{clip}(2.09 x)$ & $1d$ \\
9  & \texttt{blocks.4.norm1}  & 0.2109 & 0.2040 & $\mathrm{clip}(2.17 x)$ & $1d$ \\
10 & \texttt{blocks.4.norm2}  & 0.2330 & 0.2190 & $\mathrm{clip}(2.15 x)$ & $1d$ \\
11 & \texttt{blocks.5.norm1}  & 0.2125 & 0.2165 & $\mathrm{clip}(2.23 x)$ & $1d$ \\
12 & \texttt{blocks.5.norm2}  & 0.2050 & 0.2659 & $\mathrm{clip}(2.06 x)$ & $1d$ \\
13 & \texttt{blocks.6.norm1}  & 0.2021 & 0.2071 & $\mathrm{clip}(2.07 x)$ & $1d$ \\
14 & \texttt{blocks.6.norm2}  & 0.2108 & 0.2012 & $\mathrm{clip}(1.9 x)$ & $1d$ \\
15 & \texttt{blocks.7.norm1}  & 0.2177 & 0.2132 & $\mathrm{clip}(1.84 x)$ & $1d$ \\
16 & \texttt{blocks.7.norm2}  & 0.1530 & 0.1376 & $0.545 \cdot \mathrm{clip}(2.98 x)$ & $2d$ \\
17 & \texttt{blocks.8.norm1}  & 0.1982 & 0.2306 & $\mathrm{clip}(1.54 x)$ & $1d$ \\
18 & \texttt{blocks.8.norm2}  & 0.1989 & 0.2071 & $\mathrm{clip}(-1.38 \cdot (-x))$ & $1d$ \\
19 & \texttt{blocks.9.norm1}  & 0.0788 & 0.1030 & $0.525 \cdot \mathrm{clip}(2.42 x + 0.409)$ & $3d$ \\
20 & \texttt{blocks.9.norm2}  & 0.1751 & 0.1673 & $\mathrm{clip}(x) + 0.161$ & $1d$ \\
21 & \texttt{blocks.10.norm1} & 0.0920 & 0.1073 & $0.486 \cdot \mathrm{clip}(2 x) + 0.171$ & $3d$ \\
22 & \texttt{blocks.10.norm2} & 0.0798 & 0.1538 & $-0.0237 x + \mathrm{clip}(x)$ & $2d$ \\
23 & \texttt{blocks.11.norm1} & 0.1443 & 0.1356 & $0.542 \cdot \mathrm{clip}(x)$ & $1d$ \\
24 & \texttt{blocks.11.norm2} & 0.2714 & 0.2260 & $0.455 \cdot \mathrm{clip}(x)$ & $1d$ \\
25 & \texttt{norm}           & 0.3560 & 0.6113 & $0.441 \cdot \mathrm{clip}(x)$ & $1d$ \\
\midrule
   & \textbf{Total}           &        &        &   & $\mathbf{31d}$ \\
\bottomrule
\end{tabular}
}
\end{table}

\clearpage
\subsection{ViT-L}

Figure~\ref{fig:all49_mappings_vitl} provides the analogous visualization for the 49 normalization layers of ViT-L, mirroring the ViT-B figure above, with the corresponding symbolic expressions listed in Table~\ref{tab:gp_solutions_vitl}.

\begin{figure}[htbp]
    \centering
    \includegraphics[width=\linewidth]{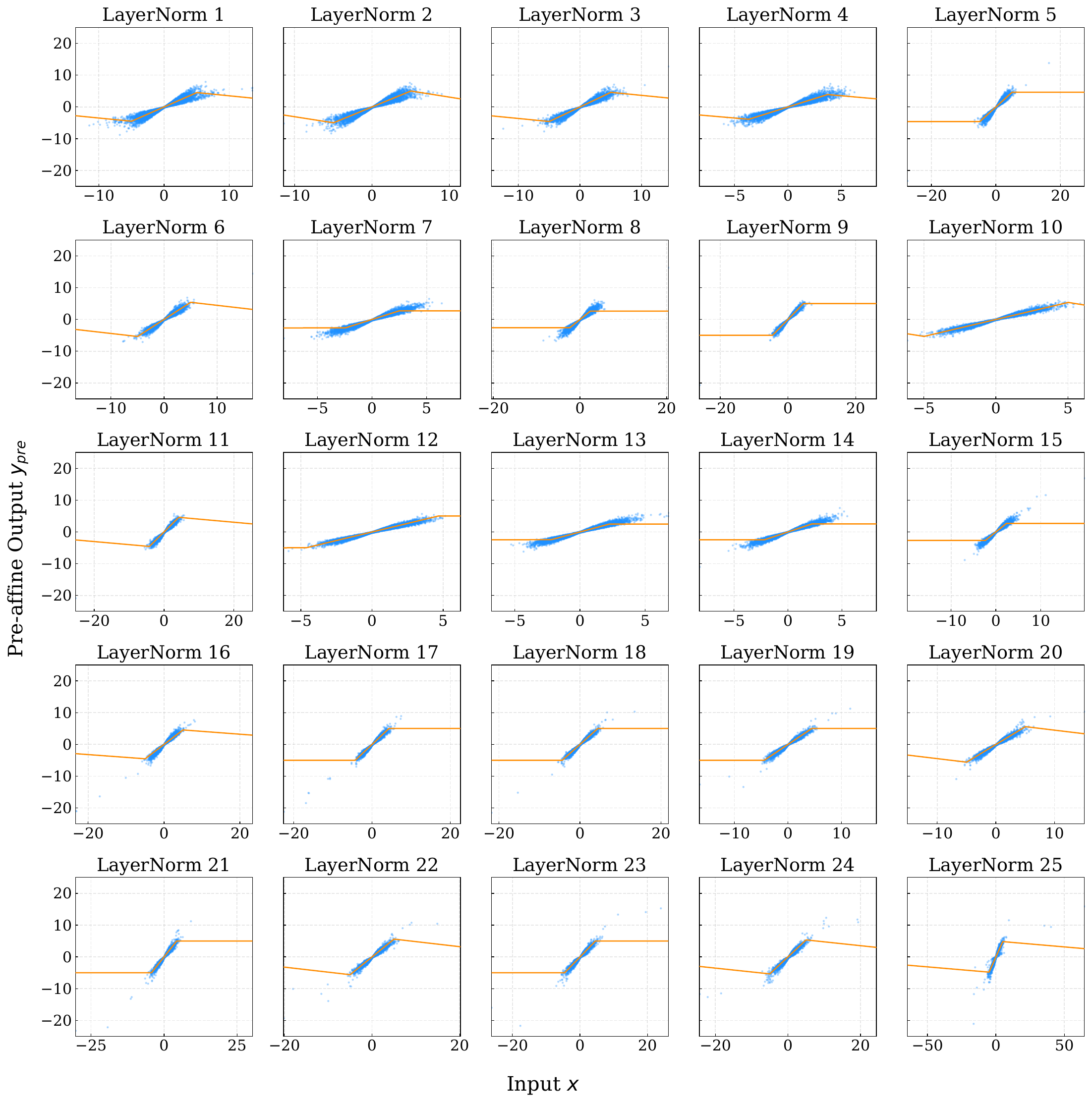}
    \caption{\textbf{Comprehensive functional alignment across all 49 normalization layers of ViT-L (1/2).} The evolved GP solutions (orange lines) are overlaid onto the target 50{,}000-point LayerNorm mappings (blue scatter) extracted from the pre-trained ViT-L architecture, mirroring Figure~\ref{fig:all25_mappings}.}
    \label{fig:all49_mappings_vitl}
\end{figure}

\begin{figure}[htbp]
    \ContinuedFloat
    \centering
    \includegraphics[width=\linewidth]{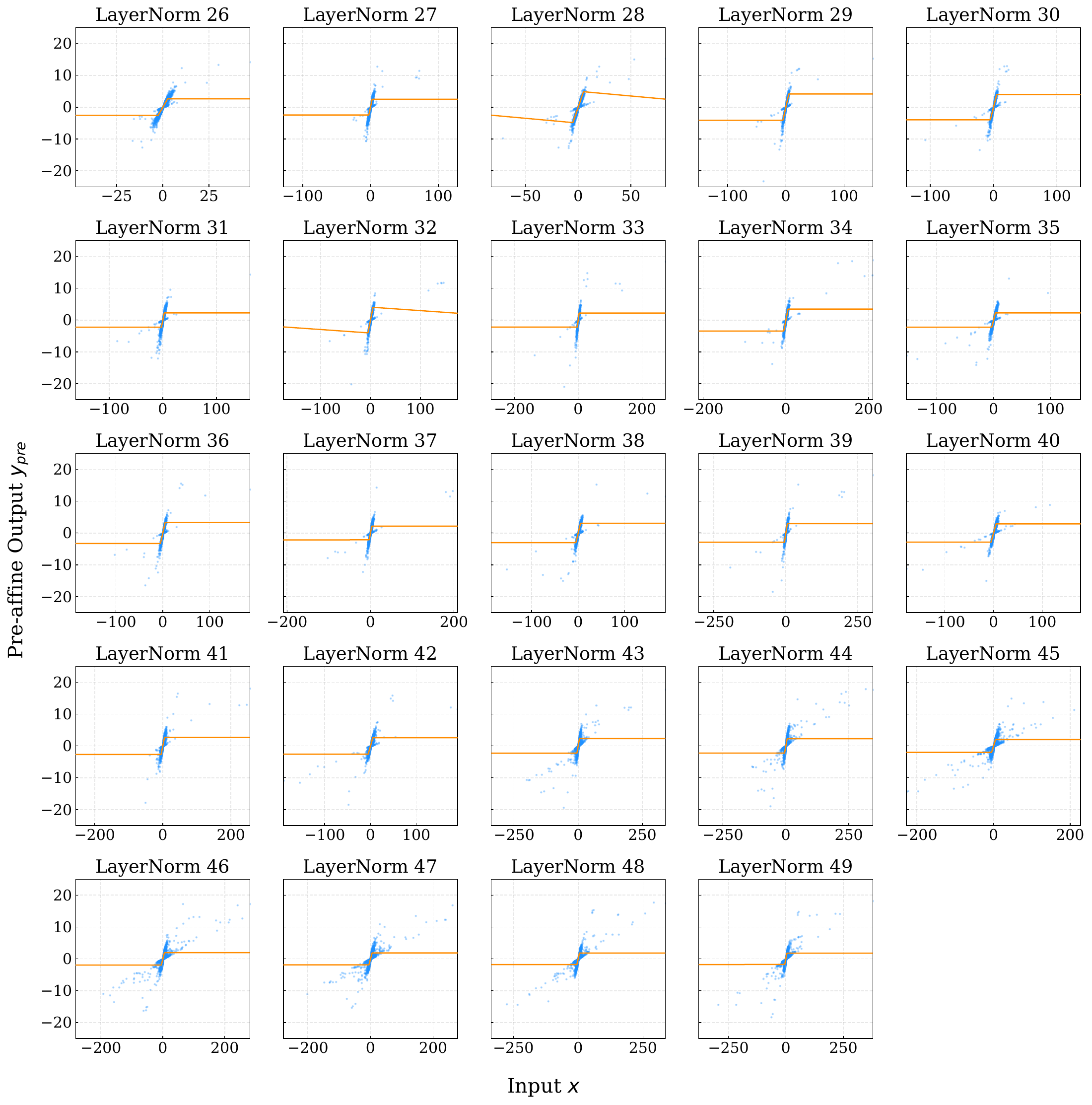}
    \caption{\textbf{(continued)} Remaining normalization layers of ViT-L.}
    \label{fig:all49_mappings_vitl_2}
\end{figure}

\clearpage
\begin{table}[htbp]
\centering
\caption{\textbf{Optimal GP-evolved symbolic expressions per layer (ViT-L).} Definitions and notation follow Table~\ref{tab:gp_solutions_vitb}.}
\label{tab:gp_solutions_vitl}
\resizebox{\textwidth}{!}{
\begin{tabular}{llccll}
\toprule
\# & Layer & Train Fitness ($\downarrow$) & Val Fitness ($\downarrow$) & Expression & FLOPs\\
\midrule
1 & \texttt{blocks.0.norm1} & 0.0696 & 0.0695 & $-0.205 x + 1.11 \cdot \mathrm{clip}(x)$ & $3d$ \\
2 & \texttt{blocks.0.norm2} & 0.0657 & 0.0862 & $-0.386 x + 1.39 \cdot \mathrm{clip}(x)$ & $3d$ \\
3 & \texttt{blocks.1.norm1} & 0.0448 & 0.0490 & $-0.191 x + 1.11 \cdot \mathrm{clip}(x)$ & $3d$ \\
4 & \texttt{blocks.1.norm2} & 0.0513 & 0.0735 & $-0.297 x + \mathrm{clip}(1.36 x)$ & $3d$ \\
5 & \texttt{blocks.2.norm1} & 0.1659 & 0.1598 & $0.926 \cdot \mathrm{clip}(x)$ & $1d$ \\
6 & \texttt{blocks.2.norm2} & 0.0332 & 0.1022 & $-0.192 x + 1.27 \cdot \mathrm{clip}(x)$ & $3d$ \\
7 & \texttt{blocks.3.norm1} & 0.0773 & 0.0771 & $0.536 \cdot \mathrm{clip}(2.04 x)$ & $2d$ \\
8 & \texttt{blocks.3.norm2} & 0.0837 & 0.0852 & $0.525 \cdot \mathrm{clip}(2.12 x)$ & $2d$ \\
9 & \texttt{blocks.4.norm1} & 0.1533 & 0.1485 & $\mathrm{clip}(-1.1 \cdot (-x))$ & $1d$ \\
10 & \texttt{blocks.4.norm2} & 0.0233 & 0.0225 & $-0.737 x + 1.81 \cdot \mathrm{clip}(x)$ & $3d$ \\
11 & \texttt{blocks.5.norm1} & 0.0305 & 0.1004 & $-0.0979 x + \mathrm{clip}(1.19 x)$ & $3d$ \\
12 & \texttt{blocks.5.norm2} & 0.1472 & 0.1468 & $\mathrm{clip}(1.08 x)$ & $1d$ \\
13 & \texttt{blocks.6.norm1} & 0.0673 & 0.0693 & $0.496 \cdot \mathrm{clip}(2.24 x)$ & $2d$ \\
14 & \texttt{blocks.6.norm2} & 0.0692 & 0.0639 & $0.499 \cdot \mathrm{clip}(2.29 x)$ & $2d$ \\
15 & \texttt{blocks.7.norm1} & 0.0778 & 0.1092 & $0.536 \cdot \mathrm{clip}(2.13 x)$ & $2d$ \\
16 & \texttt{blocks.7.norm2} & 0.0815 & 0.1414 & $-0.0896 x + \mathrm{clip}(x)$ & $2d$ \\
17 & \texttt{blocks.8.norm1} & 0.1644 & 0.1483 & $\mathrm{clip}(1.17 x)$ & $1d$ \\
18 & \texttt{blocks.8.norm2} & 0.1546 & 0.1666 & $\mathrm{clip}(1.14 x)$ & $1d$ \\
19 & \texttt{blocks.9.norm1} & 0.1505 & 0.1583 & $\mathrm{clip}(1.13 x)$ & $1d$ \\
20 & \texttt{blocks.9.norm2} & 0.0241 & 0.0969 & $-0.222 x + 1.34 \cdot \mathrm{clip}(x)$ & $3d$ \\
21 & \texttt{blocks.10.norm1} & 0.1524 & 0.2704 & $\mathrm{clip}(1.16 x)$ & $1d$ \\
22 & \texttt{blocks.10.norm2} & 0.0346 & 0.0950 & $-0.159 x + 1.28 \cdot \mathrm{clip}(x)$ & $3d$ \\
23 & \texttt{blocks.11.norm1} & 0.1655 & 0.1526 & $\mathrm{clip}(1.1 x)$ & $1d$ \\
24 & \texttt{blocks.11.norm2} & 0.0329 & 0.0663 & $-0.121 x + 1.19 \cdot \mathrm{clip}(x)$ & $3d$ \\
25 & \texttt{blocks.12.norm1} & 0.0538 & 0.0774 & $-0.0371 x + \mathrm{clip}(x)$ & $2d$ \\
26 & \texttt{blocks.12.norm2} & 0.0976 & 0.0928 & $0.519 \cdot \mathrm{clip}(1.96 x)$ & $2d$ \\
27 & \texttt{blocks.13.norm1} & 0.1059 & 0.1158 & $0.496 \cdot \mathrm{clip}(2.01 x)$ & $2d$ \\
28 & \texttt{blocks.13.norm2} & 0.0668 & 0.1072 & $-0.03 x + \mathrm{clip}(x)$ & $2d$ \\
29 & \texttt{blocks.14.norm1} & 0.1728 & 0.1763 & $0.824 \cdot \mathrm{clip}(x)$ & $1d$ \\
30 & \texttt{blocks.14.norm2} & 0.1616 & 0.1558 & $0.796 \cdot \mathrm{clip}(x)$ & $1d$ \\
31 & \texttt{blocks.15.norm1} & 0.1105 & 0.1291 & $0.45 \cdot \mathrm{clip}(1.93 x)$ & $2d$ \\
32 & \texttt{blocks.15.norm2} & 0.0790 & 0.0914 & $-0.011 x + 0.813 \cdot \mathrm{clip}(x)$ & $3d$ \\
33 & \texttt{blocks.16.norm1} & 0.1132 & 0.1896 & $0.442 \cdot \mathrm{clip}(1.78 x)$ & $2d$ \\
34 & \texttt{blocks.16.norm2} & 0.1389 & 0.1339 & $0.692 \cdot \mathrm{clip}(x)$ & $1d$ \\
35 & \texttt{blocks.17.norm1} & 0.1028 & 0.1153 & $0.448 \cdot \mathrm{clip}(1.62 x)$ & $2d$ \\
36 & \texttt{blocks.17.norm2} & 0.1279 & 0.1390 & $0.66 \cdot \mathrm{clip}(x)$ & $1d$ \\
37 & \texttt{blocks.18.norm1} & 0.1008 & 0.1057 & $0.435 \cdot \mathrm{clip}(1.57 x)$ & $2d$ \\
38 & \texttt{blocks.18.norm2} & 0.1125 & 0.1867 & $0.608 \cdot \mathrm{clip}(x)$ & $1d$ \\
39 & \texttt{blocks.19.norm1} & 0.1297 & 0.1083 & $0.584 \cdot \mathrm{clip}(x)$ & $1d$ \\
40 & \texttt{blocks.19.norm2} & 0.1011 & 0.1109 & $0.575 \cdot \mathrm{clip}(x)$ & $1d$ \\
41 & \texttt{blocks.20.norm1} & 0.1243 & 0.2006 & $0.534 \cdot \mathrm{clip}(x)$ & $1d$ \\
42 & \texttt{blocks.20.norm2} & 0.1121 & 0.1720 & $0.518 \cdot \mathrm{clip}(x)$ & $1d$ \\
43 & \texttt{blocks.21.norm1} & 0.1881 & 0.2277 & $0.459 \cdot \mathrm{clip}(x)$ & $1d$ \\
44 & \texttt{blocks.21.norm2} & 0.2204 & 0.1585 & $0.451 \cdot \mathrm{clip}(x)$ & $1d$ \\
45 & \texttt{blocks.22.norm1} & 0.1838 & 0.2024 & $0.401 \cdot \mathrm{clip}(x)$ & $1d$ \\
46 & \texttt{blocks.22.norm2} & 0.2229 & 0.1768 & $-0.391 \cdot (-\mathrm{clip}(x))$ & $1d$ \\
47 & \texttt{blocks.23.norm1} & 0.1905 & 0.1588 & $0.375 \cdot \mathrm{clip}(x)$ & $1d$ \\
48 & \texttt{blocks.23.norm2} & 0.1833 & 0.1258 & $-0.362 \cdot (-\mathrm{clip}(x))$ & $1d$ \\
49 & \texttt{norm} & 0.1769 & 0.2045 & $0.358 \cdot \mathrm{clip}(x)$ & $1d$ \\
\midrule
   & \textbf{Total}           &        &        &   & $\mathbf{85d}$ \\
\bottomrule
\end{tabular}
}
\end{table}

\clearpage
\subsection{Swin-B}

Figure~\ref{fig:all53_mappings_swinb} provides the analogous visualization for the 53 normalization layers of Swin-B, mirroring the ViT-B figure above, with the corresponding symbolic expressions listed in Table~\ref{tab:gp_solutions_swinb}.

\begin{figure}[htbp]
    \centering
    \includegraphics[width=\linewidth]{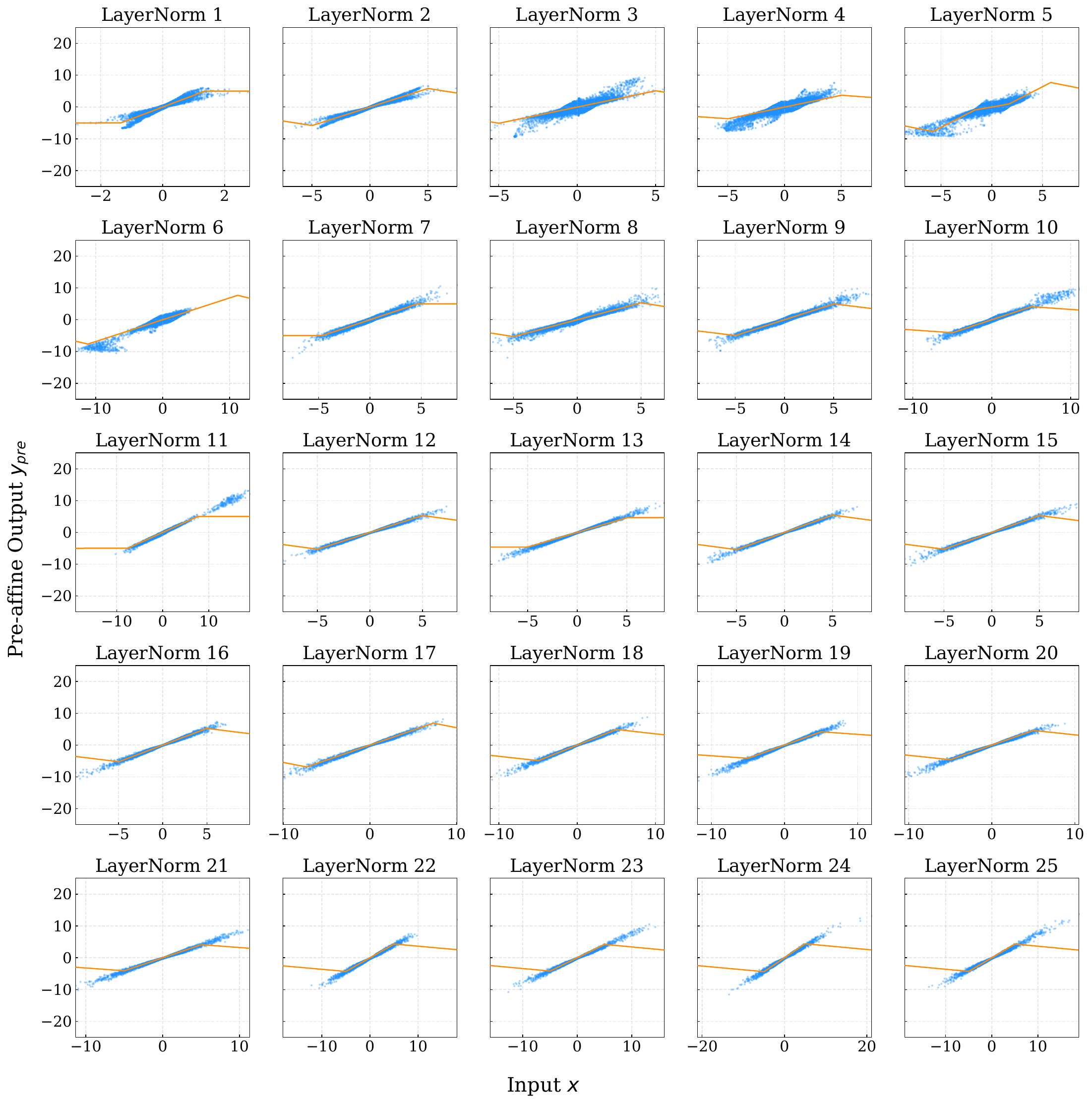}
    \caption{\textbf{Comprehensive functional alignment across all 53 normalization layers of Swin-B (1/2).} The evolved GP solutions (orange lines) are overlaid onto the target 50{,}000-point LayerNorm mappings (blue scatter) extracted from the pre-trained Swin-B architecture, mirroring Figure~\ref{fig:all25_mappings}.}
    \label{fig:all53_mappings_swinb}
\end{figure}

\begin{figure}[htbp]
    \ContinuedFloat
    \centering
    \includegraphics[width=\linewidth]{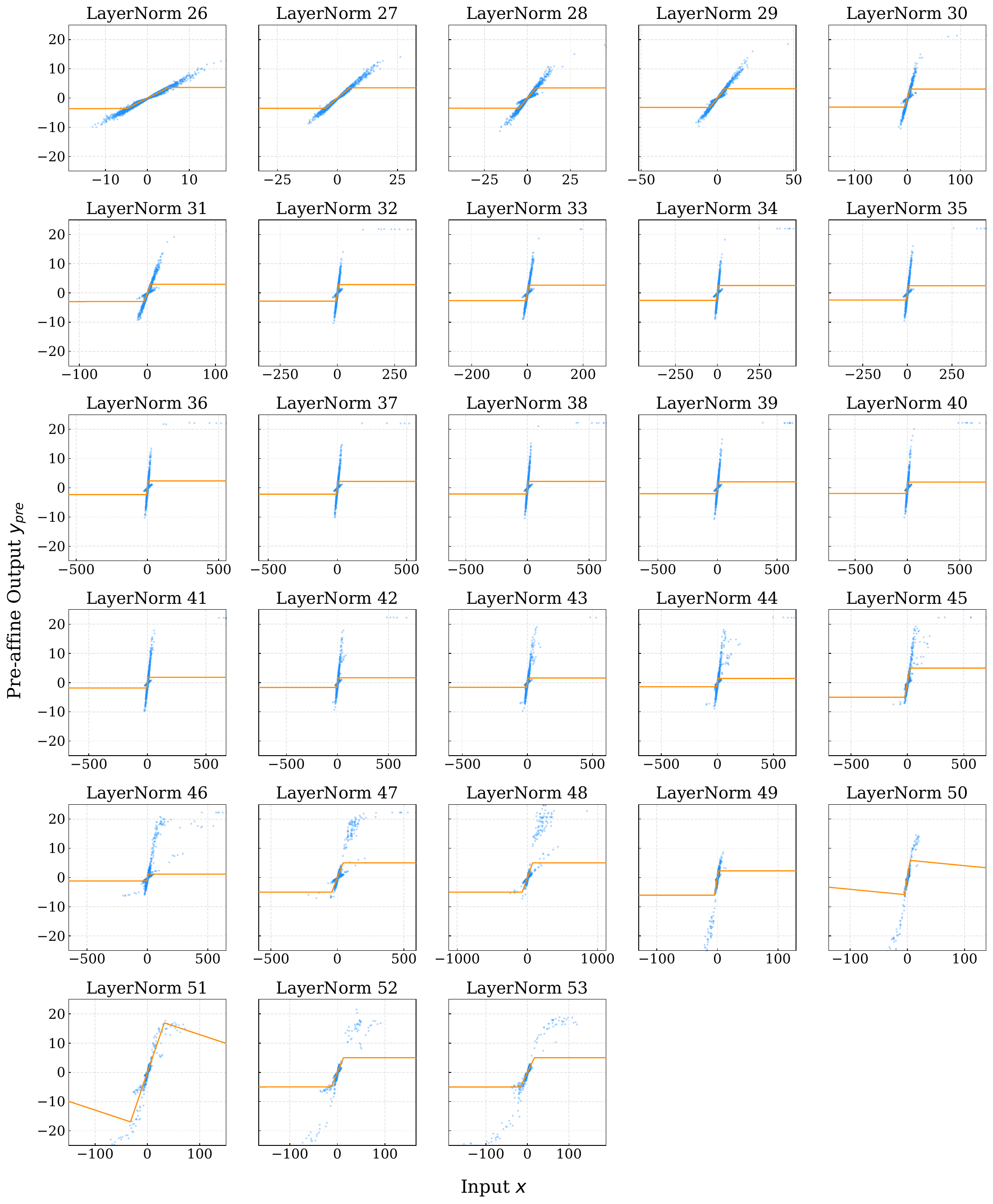}
    \caption{\textbf{(continued)} Remaining normalization layers of Swin-B.}
    \label{fig:all53_mappings_swinb_2}
\end{figure}

\begin{table}[htbp]
\centering
\caption{\textbf{Optimal GP-evolved symbolic expressions per layer (Swin-B).} Definitions and notation follow Table~\ref{tab:gp_solutions_vitb}. Here $d$ is the input width of each layer, which differs between stages, so the per-layer costs are not summed; Appendix~\ref{app:gp-flops} gives the total cost per image.}
\label{tab:gp_solutions_swinb}
\resizebox{\textwidth}{!}{
\begin{tabular}{llccll}
\toprule
\# & Layer & Train Fitness ($\downarrow$) & Val Fitness ($\downarrow$) & Expression & FLOPs\\
\midrule
1 & \texttt{patch\_embed.norm} & 0.2377 & 0.2288 & $\mathrm{clip}(3.7 x)$ & $1d$ \\
2 & \texttt{layers.0.blocks.0.norm1} & 0.0496 & 0.0574 & $-0.581 x + 1.75 \cdot \mathrm{clip}(x)$ & $3d$ \\
3 & \texttt{layers.0.blocks.0.norm2} & 0.4183 & 0.4153 & $-0.837 x + 1.86 \cdot \mathrm{clip}(x)$ & $3d$ \\
4 & \texttt{layers.0.blocks.1.norm1} & 0.3671 & 0.3844 & $-0.259 x + \mathrm{clip}(x)$ & $2d$ \\
5 & \texttt{layers.0.blocks.1.norm2} & 0.2888 & 0.2853 & $-0.635 x + 2.28 \cdot \mathrm{clip}(\mathrm{clip}(x - 0.161 \cdot \mathrm{clip}(3 x)))$ & $6d$ \\
6 & \texttt{layers.1.downsample.norm} & 0.2352 & 0.2342 & $-0.512 x + 2.68 \cdot \mathrm{clip}(\mathrm{clip}(0.447 x))$ & $4d$ \\
7 & \texttt{layers.1.blocks.0.norm1} & 0.1904 & 0.1849 & $\mathrm{clip}(1.08 x)$ & $1d$ \\
8 & \texttt{layers.1.blocks.0.norm2} & 0.0773 & 0.0735 & $-0.644 x + 1.71 \cdot \mathrm{clip}(x)$ & $3d$ \\
9 & \texttt{layers.1.blocks.1.norm1} & 0.0744 & 0.0812 & $1.38 \cdot \mathrm{clip}(x) + 0.382 \cdot (-x)$ & $3d$ \\
10 & \texttt{layers.1.blocks.1.norm2} & 0.1089 & 0.1383 & $-0.177 x + \mathrm{clip}(x)$ & $2d$ \\
11 & \texttt{layers.2.downsample.norm} & 0.2439 & 0.2456 & $\mathrm{clip}(0.658 x)$ & $1d$ \\
12 & \texttt{layers.2.blocks.0.norm1} & 0.0110 & 0.0159 & $-0.457 x + 1.52 \cdot \mathrm{clip}(x)$ & $3d$ \\
13 & \texttt{layers.2.blocks.0.norm2} & 0.1352 & 0.1369 & $0.927 \cdot \mathrm{clip}(x)$ & $1d$ \\
14 & \texttt{layers.2.blocks.1.norm1} & 0.0176 & 0.0333 & $-0.41 x + 1.5 \cdot \mathrm{clip}(x)$ & $3d$ \\
15 & \texttt{layers.2.blocks.1.norm2} & 0.0236 & 0.0375 & $-0.389 x + 1.45 \cdot \mathrm{clip}(x)$ & $3d$ \\
16 & \texttt{layers.2.blocks.2.norm1} & 0.0247 & 0.0230 & $-0.348 x + 1.4 \cdot \mathrm{clip}(x)$ & $3d$ \\
17 & \texttt{layers.2.blocks.2.norm2} & 0.0120 & 0.0198 & $2.17 \cdot \mathrm{clip}(0.682 x) + (-0.538 x)$ & $4d$ \\
18 & \texttt{layers.2.blocks.3.norm1} & 0.0308 & 0.0370 & $-0.277 x + 1.26 \cdot \mathrm{clip}(x)$ & $3d$ \\
19 & \texttt{layers.2.blocks.3.norm2} & 0.0523 & 0.0624 & $-0.163 x + \mathrm{clip}(x)$ & $2d$ \\
20 & \texttt{layers.2.blocks.4.norm1} & 0.0333 & 0.0343 & $-0.278 x + 1.2 \cdot \mathrm{clip}(x)$ & $3d$ \\
21 & \texttt{layers.2.blocks.4.norm2} & 0.0526 & 0.0619 & $-0.178 x + \mathrm{clip}(x)$ & $2d$ \\
22 & \texttt{layers.2.blocks.5.norm1} & 0.0321 & 0.0679 & $-0.136 x + \mathrm{clip}(x)$ & $2d$ \\
23 & \texttt{layers.2.blocks.5.norm2} & 0.0634 & 0.0553 & $-0.16 x + \mathrm{clip}(x)$ & $2d$ \\
24 & \texttt{layers.2.blocks.6.norm1} & 0.0484 & 0.0662 & $-0.12 x + \mathrm{clip}(x)$ & $2d$ \\
25 & \texttt{layers.2.blocks.6.norm2} & 0.0850 & 0.0814 & $-0.137 x + \mathrm{clip}(x)$ & $2d$ \\
26 & \texttt{layers.2.blocks.7.norm1} & 0.1192 & 0.1525 & $0.724 \cdot \mathrm{clip}(x)$ & $1d$ \\
27 & \texttt{layers.2.blocks.7.norm2} & 0.1380 & 0.1563 & $0.7 \cdot \mathrm{clip}(x)$ & $1d$ \\
28 & \texttt{layers.2.blocks.8.norm1} & 0.1401 & 0.1337 & $0.695 \cdot \mathrm{clip}(x)$ & $1d$ \\
29 & \texttt{layers.2.blocks.8.norm2} & 0.1513 & 0.1129 & $0.641 \cdot \mathrm{clip}(x)$ & $1d$ \\
30 & \texttt{layers.2.blocks.9.norm1} & 0.1882 & 0.1469 & $0.619 \cdot \mathrm{clip}(x)$ & $1d$ \\
31 & \texttt{layers.2.blocks.9.norm2} & 0.1853 & 0.2147 & $0.593 \cdot \mathrm{clip}(x)$ & $1d$ \\
32 & \texttt{layers.2.blocks.10.norm1} & 0.2644 & 0.2388 & $0.566 \cdot \mathrm{clip}(x)$ & $1d$ \\
33 & \texttt{layers.2.blocks.10.norm2} & 0.2198 & 0.2835 & $0.526 \cdot \mathrm{clip}(x)$ & $1d$ \\
34 & \texttt{layers.2.blocks.11.norm1} & 0.2604 & 0.4177 & $0.509 \cdot \mathrm{clip}(x)$ & $1d$ \\
35 & \texttt{layers.2.blocks.11.norm2} & 0.2800 & 0.2067 & $0.488 \cdot \mathrm{clip}(x)$ & $1d$ \\
36 & \texttt{layers.2.blocks.12.norm1} & 0.2487 & 0.4365 & $0.469 \cdot \mathrm{clip}(x)$ & $1d$ \\
37 & \texttt{layers.2.blocks.12.norm2} & 0.2579 & 0.2618 & $0.438 \cdot \mathrm{clip}(x)$ & $1d$ \\
38 & \texttt{layers.2.blocks.13.norm1} & 0.3015 & 0.1940 & $0.432 \cdot \mathrm{clip}(x)$ & $1d$ \\
39 & \texttt{layers.2.blocks.13.norm2} & 0.2906 & 0.2043 & $0.407 \cdot \mathrm{clip}(x)$ & $1d$ \\
40 & \texttt{layers.2.blocks.14.norm1} & 0.2569 & 0.3801 & $0.388 \cdot \mathrm{clip}(x)$ & $1d$ \\
41 & \texttt{layers.2.blocks.14.norm2} & 0.2733 & 0.3857 & $0.366 \cdot \mathrm{clip}(x)$ & $1d$ \\
42 & \texttt{layers.2.blocks.15.norm1} & 0.2678 & 0.1953 & $-0.332 \cdot \mathrm{clip}((-x))$ & $1d$ \\
43 & \texttt{layers.2.blocks.15.norm2} & 0.2676 & 0.1958 & $0.319 \cdot \mathrm{clip}(x)$ & $1d$ \\
44 & \texttt{layers.2.blocks.16.norm1} & 0.3326 & 0.1815 & $0.286 \cdot \mathrm{clip}(x)$ & $1d$ \\
45 & \texttt{layers.2.blocks.16.norm2} & 0.3419 & 0.2893 & $\mathrm{clip}(0.215 x)$ & $1d$ \\
46 & \texttt{layers.2.blocks.17.norm1} & 0.4776 & 0.7679 & $0.235 \cdot \mathrm{clip}(x)$ & $1d$ \\
47 & \texttt{layers.2.blocks.17.norm2} & 0.5053 & 0.5624 & $\mathrm{clip}(0.118 x)$ & $1d$ \\
48 & \texttt{layers.3.downsample.norm} & 0.6214 & 0.6481 & $\mathrm{clip}(0.0678 x)$ & $1d$ \\
49 & \texttt{layers.3.blocks.0.norm1} & 0.4802 & 0.4026 & $-0.835 \cdot (-\mathrm{clip}(1.61 x + 2.27)) - 1.88$ & $4d$ \\
50 & \texttt{layers.3.blocks.0.norm2} & 0.3737 & 0.3538 & $-0.0191 x + 1.19 \cdot \mathrm{clip}(x)$ & $3d$ \\
51 & \texttt{layers.3.blocks.1.norm1} & 0.1920 & 0.2678 & $-0.0594 x + 3.77 \cdot \mathrm{clip}(0.155 x)$ & $4d$ \\
52 & \texttt{layers.3.blocks.1.norm2} & 0.5141 & 0.5697 & $\mathrm{clip}(0.39 x)$ & $1d$ \\
53 & \texttt{norm} & 0.6052 & 0.6031 & $\mathrm{clip}(0.298 x)$ & $1d$ \\
\bottomrule
\end{tabular}
}
\end{table}

\clearpage
\subsection{ConvNeXt-B}

Figure~\ref{fig:all41_mappings_convnextb} provides the analogous visualization for the 41 normalization layers of ConvNeXt-B, mirroring the ViT-B figure above, with the corresponding symbolic expressions listed in Table~\ref{tab:gp_solutions_convnextb}.

\begin{figure}[htbp]
    \centering
    \includegraphics[width=\linewidth]{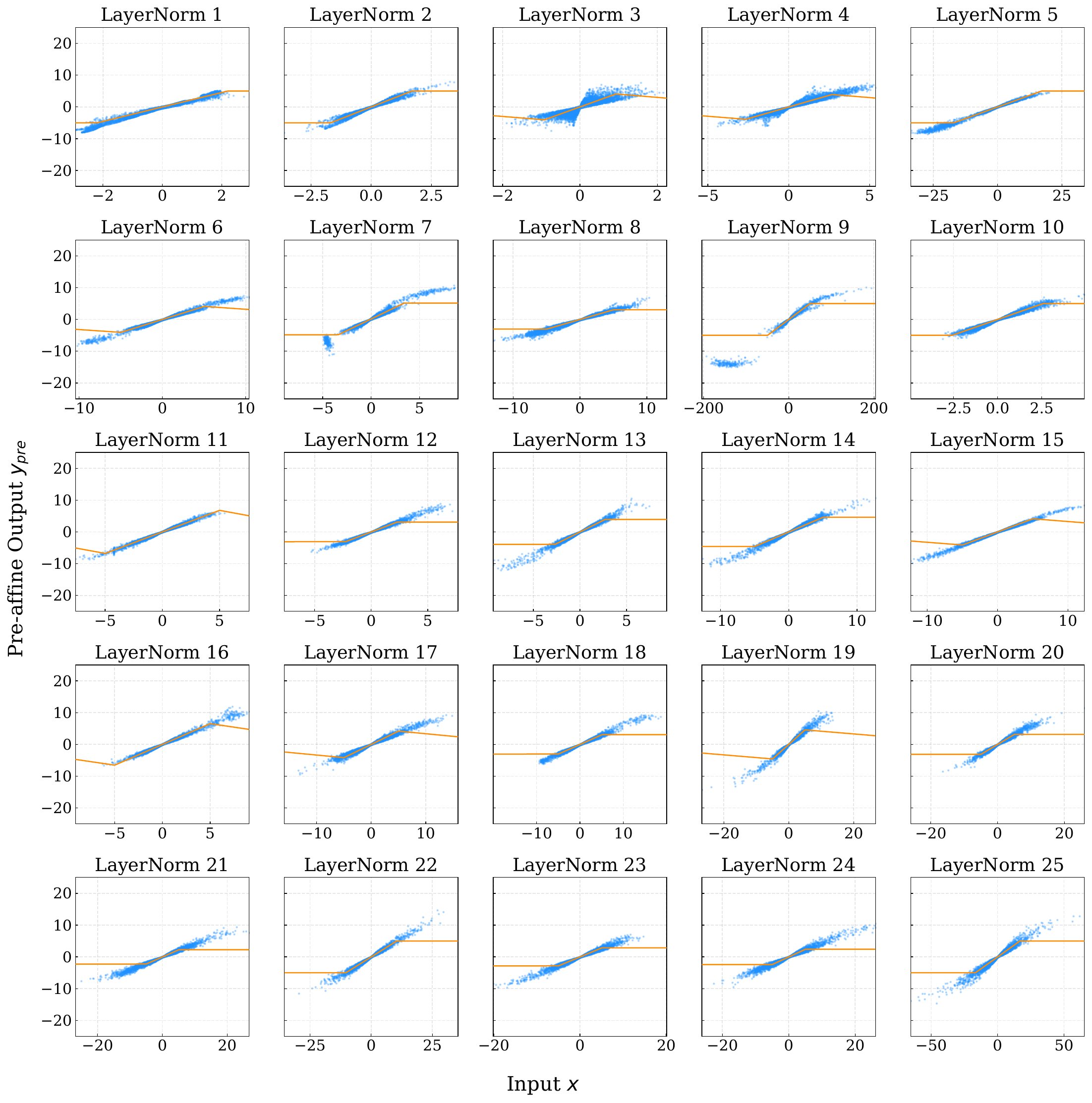}
    \caption{\textbf{Comprehensive functional alignment across all 41 normalization layers of ConvNeXt-B (1/2).} The evolved GP solutions (orange lines) are overlaid onto the target 50{,}000-point LayerNorm mappings (blue scatter) extracted from the pre-trained ConvNeXt-B architecture, mirroring Figure~\ref{fig:all25_mappings}.}
    \label{fig:all41_mappings_convnextb}
\end{figure}

\begin{figure}[htbp]
    \ContinuedFloat
    \centering
    \includegraphics[width=\linewidth]{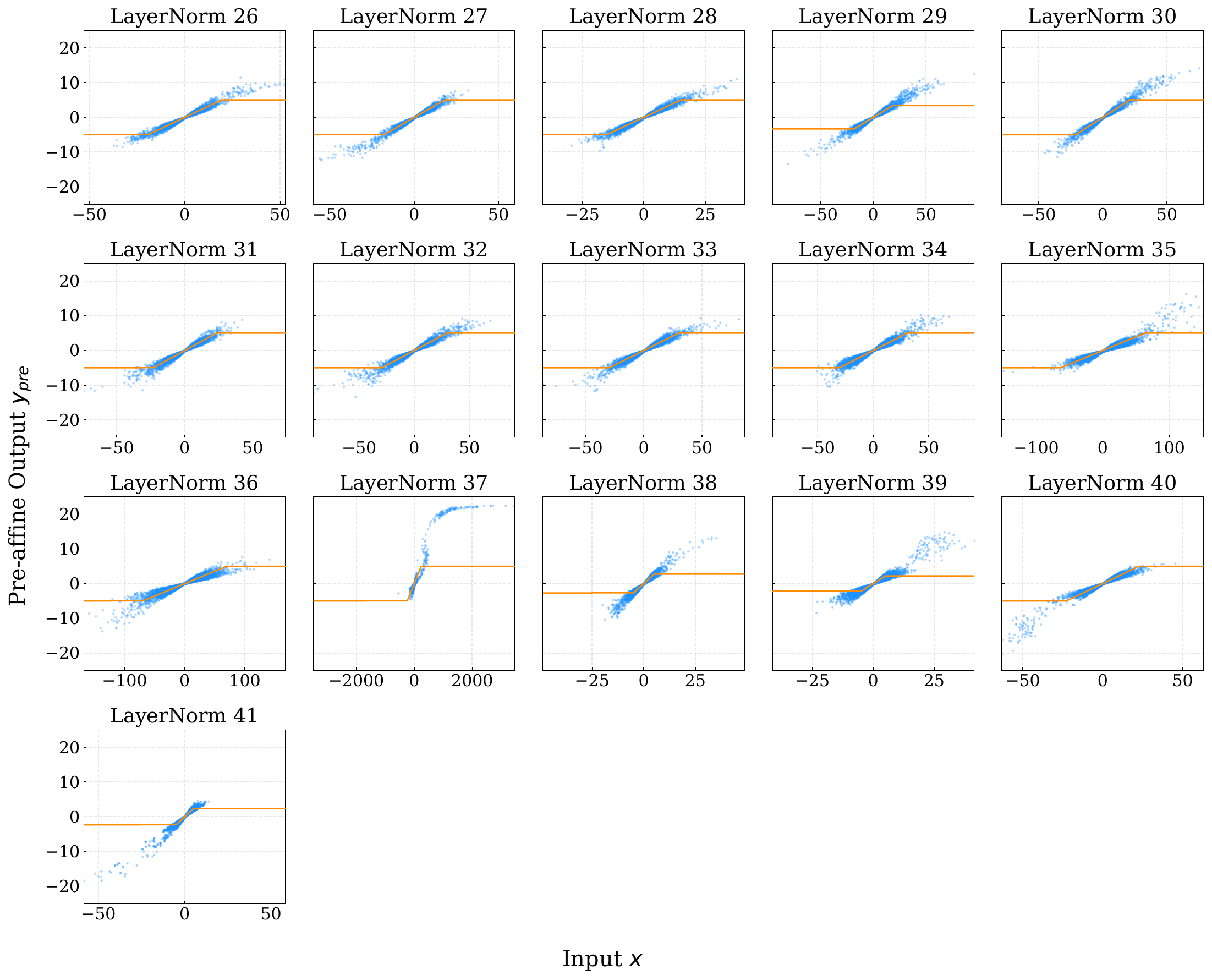}
    \caption{\textbf{(continued)} Remaining normalization layers of ConvNeXt-B.}
    \label{fig:all41_mappings_convnextb_2}
\end{figure}

\begin{table}[htbp]
\centering
\caption{\textbf{Optimal GP-evolved symbolic expressions per layer (ConvNeXt-B).} Definitions and notation follow Table~\ref{tab:gp_solutions_swinb}.}
\label{tab:gp_solutions_convnextb}
\resizebox{\textwidth}{!}{
\begin{tabular}{llccll}
\toprule
\# & Layer & Train Fitness ($\downarrow$) & Val Fitness ($\downarrow$) & Expression & FLOPs\\
\midrule
1 & \texttt{stem.1} & 0.2127 & 0.2145 & $\mathrm{clip}(2.28 x)$ & $1d$ \\
2 & \texttt{stages.0.blocks.0.norm} & 0.1643 & 0.1635 & $\mathrm{clip}(2.97 x)$ & $1d$ \\
3 & \texttt{stages.0.blocks.1.norm} & 0.2191 & 0.2378 & $\mathrm{clip}(5.41 x) + (-x)$ & $2d$ \\
4 & \texttt{stages.0.blocks.2.norm} & 0.0728 & 0.0817 & $-0.412 x + \mathrm{clip}(1.88 x)$ & $3d$ \\
5 & \texttt{stages.1.downsample.0} & 0.1674 & 0.1596 & $\mathrm{clip}(0.294 x)$ & $1d$ \\
6 & \texttt{stages.1.blocks.0.norm} & 0.0689 & 0.0562 & $-0.181 x + \mathrm{clip}(x)$ & $2d$ \\
7 & \texttt{stages.1.blocks.1.norm} & 0.2070 & 0.1977 & $\mathrm{clip}(1.47 x) + 0.159$ & $2d$ \\
8 & \texttt{stages.1.blocks.2.norm} & 0.0872 & 0.0931 & $0.611 \cdot \mathrm{clip}(x)$ & $1d$ \\
9 & \texttt{stages.2.downsample.0} & 0.4494 & 0.4619 & $\mathrm{clip}(0.101 x)$ & $1d$ \\
10 & \texttt{stages.2.blocks.0.norm} & 0.1588 & 0.1612 & $\mathrm{clip}(1.95 x)$ & $1d$ \\
11 & \texttt{stages.2.blocks.1.norm} & 0.0188 & 0.0529 & $-0.663 x + 2.02 \cdot \mathrm{clip}(x)$ & $3d$ \\
12 & \texttt{stages.2.blocks.2.norm} & 0.0945 & 0.0837 & $0.618 \cdot \mathrm{clip}(2.2 x)$ & $2d$ \\
13 & \texttt{stages.2.blocks.3.norm} & 0.1636 & 0.1312 & $0.786 \cdot \mathrm{clip}(1.87 x)$ & $2d$ \\
14 & \texttt{stages.2.blocks.4.norm} & 0.1595 & 0.1763 & $0.92 \cdot \mathrm{clip}(x)$ & $1d$ \\
15 & \texttt{stages.2.blocks.5.norm} & 0.0621 & 0.0706 & $-0.173 x + \mathrm{clip}(x)$ & $2d$ \\
16 & \texttt{stages.2.blocks.6.norm} & 0.0539 & 0.0538 & $-0.443 x + 1.75 \cdot \mathrm{clip}(x)$ & $3d$ \\
17 & \texttt{stages.2.blocks.7.norm} & 0.0667 & 0.0861 & $-0.164 x + \mathrm{clip}(x)$ & $2d$ \\
18 & \texttt{stages.2.blocks.8.norm} & 0.1171 & 0.1116 & $0.61 \cdot \mathrm{clip}(x)$ & $1d$ \\
19 & \texttt{stages.2.blocks.9.norm} & 0.0949 & 0.0958 & $-0.0852 x + \mathrm{clip}(x)$ & $2d$ \\
20 & \texttt{stages.2.blocks.10.norm} & 0.1084 & 0.1113 & $0.629 \cdot \mathrm{clip}(x)$ & $1d$ \\
21 & \texttt{stages.2.blocks.11.norm} & 0.1315 & 0.1291 & $0.451 \cdot \mathrm{clip}(x)$ & $1d$ \\
22 & \texttt{stages.2.blocks.12.norm} & 0.1742 & 0.1693 & $\mathrm{clip}(0.483 x)$ & $1d$ \\
23 & \texttt{stages.2.blocks.13.norm} & 0.1112 & 0.1250 & $0.567 \cdot \mathrm{clip}(x)$ & $1d$ \\
24 & \texttt{stages.2.blocks.14.norm} & 0.1434 & 0.1611 & $0.485 \cdot \mathrm{clip}(x)$ & $1d$ \\
25 & \texttt{stages.2.blocks.15.norm} & 0.2062 & 0.2126 & $\mathrm{clip}(0.3 x)$ & $1d$ \\
26 & \texttt{stages.2.blocks.16.norm} & 0.1682 & 0.1759 & $\mathrm{clip}(0.26 x)$ & $1d$ \\
27 & \texttt{stages.2.blocks.17.norm} & 0.1750 & 0.1730 & $\mathrm{clip}(0.28 x)$ & $1d$ \\
28 & \texttt{stages.2.blocks.18.norm} & 0.1601 & 0.1608 & $\mathrm{clip}(0.332 x)$ & $1d$ \\
29 & \texttt{stages.2.blocks.19.norm} & 0.1450 & 0.1456 & $0.673 \cdot \mathrm{clip}(0.276 x)$ & $2d$ \\
30 & \texttt{stages.2.blocks.20.norm} & 0.1995 & 0.2095 & $\mathrm{clip}(-0.24 \cdot (-x))$ & $1d$ \\
31 & \texttt{stages.2.blocks.21.norm} & 0.1700 & 0.1705 & $\mathrm{clip}(0.217 x)$ & $1d$ \\
32 & \texttt{stages.2.blocks.22.norm} & 0.1801 & 0.1674 & $\mathrm{clip}(0.169 x)$ & $1d$ \\
33 & \texttt{stages.2.blocks.23.norm} & 0.1692 & 0.1772 & $\mathrm{clip}(0.169 x)$ & $1d$ \\
34 & \texttt{stages.2.blocks.24.norm} & 0.1834 & 0.1830 & $\mathrm{clip}(0.164 x)$ & $1d$ \\
35 & \texttt{stages.2.blocks.25.norm} & 0.2069 & 0.2426 & $\mathrm{clip}(0.0792 x)$ & $1d$ \\
36 & \texttt{stages.2.blocks.26.norm} & 0.2078 & 0.2145 & $\mathrm{clip}(0.072 x)$ & $1d$ \\
37 & \texttt{stages.3.downsample.0} & 0.7139 & 0.4462 & $\mathrm{clip}(0.0207 x)$ & $1d$ \\
38 & \texttt{stages.3.blocks.0.norm} & 0.1487 & 0.1709 & $0.541 \cdot \mathrm{clip}(x)$ & $1d$ \\
39 & \texttt{stages.3.blocks.1.norm} & 0.2305 & 0.1433 & $0.44 \cdot \mathrm{clip}(x)$ & $1d$ \\
40 & \texttt{stages.3.blocks.2.norm} & 0.2501 & 0.3082 & $\mathrm{clip}(0.22 x)$ & $1d$ \\
41 & \texttt{head.norm} & 0.2090 & 0.1149 & $0.47 \cdot \mathrm{clip}(x)$ & $1d$ \\
\bottomrule
\end{tabular}
}
\end{table}

\clearpage
\section{Fine-tuning results}
\subsection{Complete fine-tuning trajectories}
\label{A:finetune}
\begin{figure}[htbp]
    \centering
    \includegraphics[width=\textwidth]{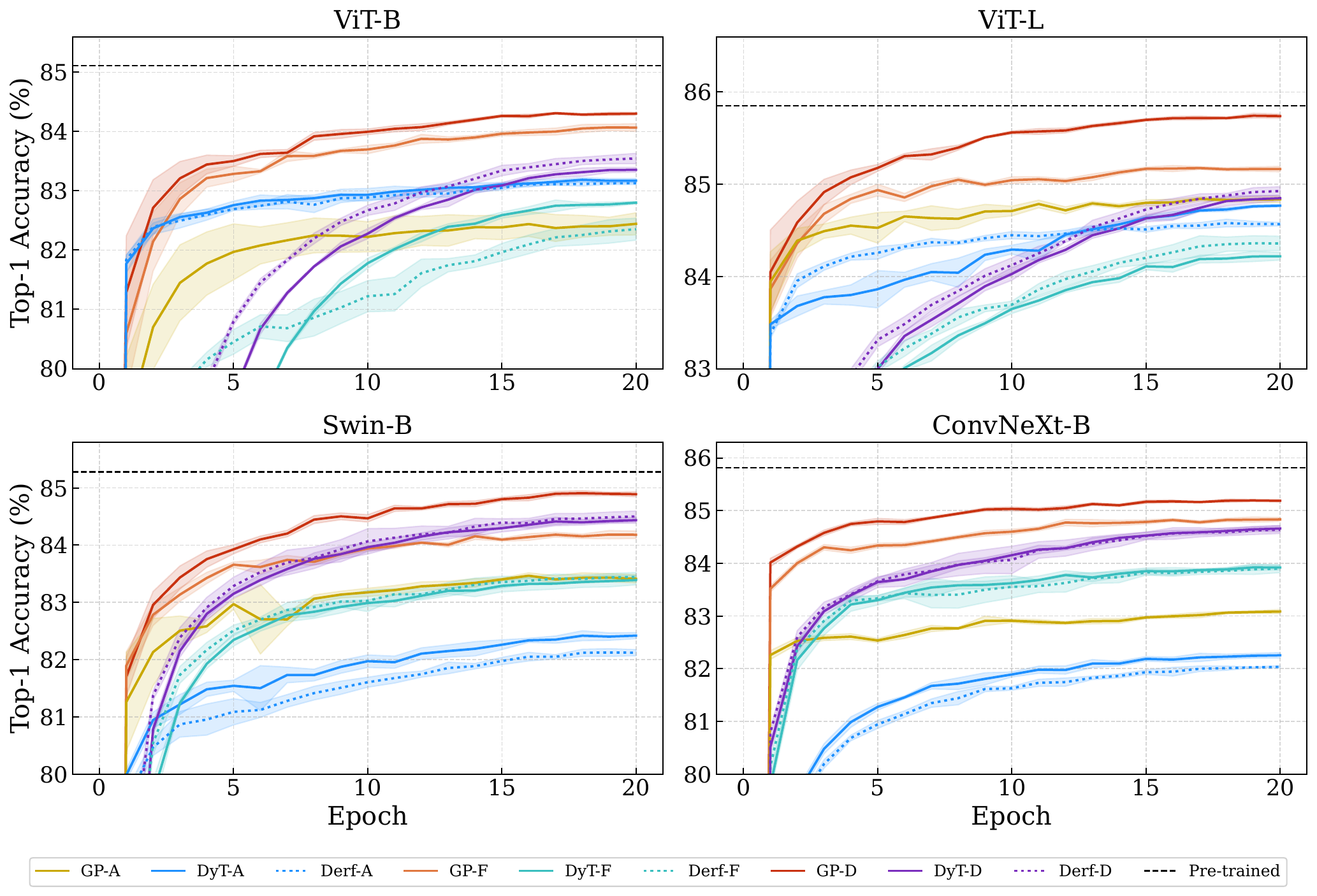}
    \caption{\textbf{ImageNet-1K validation performance recovery dynamics.} Fine-tuning trajectories of GP-A, GP-F, GP-D, DyT-A, DyT-F, and DyT-D, and of Derf-A, Derf-F, and Derf-D (dotted, in the colour of the corresponding DyT variant), at ViT-B (top left), ViT-L (top right), Swin-B (bottom left), and ConvNeXt-B (bottom right). Shaded areas represent $\pm 1$ standard deviation across five independent seeds. Black dashed lines show the accuracy of the original pre-trained model ($85.11\%$, $85.85\%$, $85.28\%$, and $85.81\%$, respectively).}
    \label{fig:full_accuracy_curves}
\end{figure}

As noted in Section~\ref{recovery}, the full fine-tuning (DyT-F) and distillation (DyT-D) variants of the Dynamic Tanh baseline converged more slowly than our GP solutions at ViT-B and ViT-L. Figure~\ref{fig:full_accuracy_curves} shows the complete trajectories of all nine variants, including Derf, for all four architectures. At ViT-B and ViT-L, both DyT variants recover at a much slower rate and level off below GP-F and GP-D. At Swin-B and ConvNeXt-B, DyT-F and DyT-D recover much faster than at the ViT models, but they still finish below their GP counterparts (GP-F and GP-D, respectively) on both architectures. Derf closely follows DyT in every variant. At ViT-B and ViT-L, Derf-F and Derf-D recover faster than DyT-F and DyT-D in the first epochs, but they stay below GP-F and GP-D throughout, as on the other two architectures.

\subsection{Robustness to distribution shift}
\label{A:robustness}
To test whether the recovered accuracy also holds outside the ImageNet-1K validation set, we evaluate every fine-tuned model from Table~\ref{tab:main_results} on four test sets with a distribution shift: ImageNet-V2 (matched frequency)~\citep{recht2019imagenet}, ImageNet-Sketch~\citep{wang2019learning}, ImageNet-R~\citep{hendrycks2021many}, and ImageNet-A~\citep{hendrycks2021natural}. ImageNet-R and ImageNet-A cover 200 of the 1000 classes, so for these two sets we only use the logits of those classes. For each run we use the checkpoint with the highest validation accuracy (the one reported in Table~\ref{tab:main_results}), and the LN row is again the original pre-trained model without fine-tuning. The results are shown in Table~\ref{tab:robustness}.

GP-F and GP-D outperform DyT-F and DyT-D on every test set at all four architectures, and the difference grows as the shift gets harder: on ImageNet-A, GP-D is $4.0$ to $8.9$ points above DyT-D. ImageNet-V2 was not used for any hyperparameter choice or checkpoint selection, so the lead of GP-D there ($0.9$ to $1.6$ points) is not a result of tuning on the validation set. With affine-only fine-tuning, GP-A also leads on every test set at Swin-B and ConvNeXt-B. At ViT-B, however, it trails DyT-A on every set (in line with Table~\ref{tab:main_results}), and at ViT-L the results are mixed. Finally, on ImageNet-A all re-aligned models stay below the pre-trained model, GP-D by $1.8$ to $10.4$ points and DyT-D by $10.3$ to $15.8$ points.

\begin{table}[htbp]
  \small
  \centering
\caption{\textbf{Accuracy under distribution shift.} Top-1 accuracy (\%) on ImageNet-V2, ImageNet-Sketch, ImageNet-R, and ImageNet-A of the fine-tuned models in Table~\ref{tab:main_results} (best-validation checkpoint, mean over 5 runs). The standard deviation is at most $0.43$, except for GP-A (up to $0.70$ at Swin-B and $1.06$ at ViT-B). The LN row is the original pre-trained model without fine-tuning. Bold marks the better of DyT and GP for each test set.}
  \label{tab:robustness}
  \setlength{\tabcolsep}{4pt}
  \begin{tabular}{llcccccccc}
    \toprule
     & & \multicolumn{2}{c}{\textbf{V2}} & \multicolumn{2}{c}{\textbf{Sketch}} & \multicolumn{2}{c}{\textbf{R}} & \multicolumn{2}{c}{\textbf{A}} \\
    \cmidrule(lr){3-4} \cmidrule(lr){5-6} \cmidrule(lr){7-8} \cmidrule(lr){9-10}
    \textbf{Model} & \textbf{Variant} & DyT & GP & DyT & GP & DyT & GP & DyT & GP \\
    \midrule
    ViT-B & LN (Pre-trained) & \multicolumn{2}{c}{$75.28$} & \multicolumn{2}{c}{$45.13$} & \multicolumn{2}{c}{$59.23$} & \multicolumn{2}{c}{$49.64$} \\
     & Affine-only & $\mathbf{72.65}$ & $71.75$ & $\mathbf{37.10}$ & $34.53$ & $\mathbf{52.50}$ & $51.10$ & $\mathbf{35.84}$ & $32.62$ \\
     & Full & $72.00$ & $\mathbf{73.73}$ & $36.14$ & $\mathbf{40.77}$ & $50.10$ & $\mathbf{54.93}$ & $33.08$ & $\mathbf{39.64}$ \\
     & Distillation & $72.75$ & $\mathbf{74.19}$ & $37.88$ & $\mathbf{40.92}$ & $52.78$ & $\mathbf{55.40}$ & $33.84$ & $\mathbf{39.27}$ \\
    \midrule
    ViT-L & LN (Pre-trained) & \multicolumn{2}{c}{$76.29$} & \multicolumn{2}{c}{$51.82$} & \multicolumn{2}{c}{$64.35$} & \multicolumn{2}{c}{$55.29$} \\
     & Affine-only & $\mathbf{74.88}$ & $74.77$ & $\mathbf{47.80}$ & $47.05$ & $61.94$ & $\mathbf{62.47}$ & $51.63$ & $\mathbf{52.37}$ \\
     & Full & $74.28$ & $\mathbf{75.55}$ & $43.81$ & $\mathbf{48.57}$ & $56.98$ & $\mathbf{62.58}$ & $44.38$ & $\mathbf{53.30}$ \\
     & Distillation & $74.92$ & $\mathbf{76.53}$ & $46.13$ & $\mathbf{50.73}$ & $58.99$ & $\mathbf{63.69}$ & $44.60$ & $\mathbf{53.45}$ \\
    \midrule
    Swin-B & LN (Pre-trained) & \multicolumn{2}{c}{$75.70$} & \multicolumn{2}{c}{$45.58$} & \multicolumn{2}{c}{$59.33$} & \multicolumn{2}{c}{$51.69$} \\
     & Affine-only & $71.67$ & $\mathbf{72.88}$ & $38.40$ & $\mathbf{39.92}$ & $51.90$ & $\mathbf{53.73}$ & $33.61$ & $\mathbf{38.22}$ \\
     & Full & $72.90$ & $\mathbf{74.21}$ & $39.61$ & $\mathbf{42.04}$ & $53.84$ & $\mathbf{56.37}$ & $37.45$ & $\mathbf{42.41}$ \\
     & Distillation & $74.23$ & $\mathbf{75.23}$ & $41.90$ & $\mathbf{43.55}$ & $55.72$ & $\mathbf{57.88}$ & $41.35$ & $\mathbf{45.36}$ \\
    \midrule
    ConvNeXt-B & LN (Pre-trained) & \multicolumn{2}{c}{$76.12$} & \multicolumn{2}{c}{$48.79$} & \multicolumn{2}{c}{$61.78$} & \multicolumn{2}{c}{$54.73$} \\
     & Affine-only & $71.46$ & $\mathbf{72.55}$ & $37.82$ & $\mathbf{40.62}$ & $51.94$ & $\mathbf{53.82}$ & $29.49$ & $\mathbf{34.54}$ \\
     & Full & $73.62$ & $\mathbf{74.97}$ & $42.94$ & $\mathbf{45.29}$ & $55.86$ & $\mathbf{58.47}$ & $38.78$ & $\mathbf{43.82}$ \\
     & Distillation & $74.33$ & $\mathbf{75.23}$ & $44.57$ & $\mathbf{46.38}$ & $57.19$ & $\mathbf{59.23}$ & $42.09$ & $\mathbf{46.20}$ \\
    \bottomrule
  \end{tabular}
\end{table}

\section{Hardware-centric cost analysis}
\label{A:Hardware-centric}

This appendix derives the worst-case FLOP cost of all normalization operations referenced in Section~\ref{sec:complexity}, under IEEE-754 FP32 unit-roundoff accuracy $u = 2^{-24} \approx 5.96 \times 10^{-8}$. We adopt the convention that each multiplication, addition, subtraction, division, and floating-point round-to-integer counts as exactly one FLOP. Transcendental functions are decomposed into a single unified primitive, the exponential, so that the FLOP bound for $\exp$ propagates to the sigmoid, the hyperbolic tangent, and the error function. The resulting per-token costs underpin the per-layer breakdown in Figure~\ref{fig:complexity} (Left) and the aggregate comparison in Figure~\ref{fig:complexity} (Right). Appendix~\ref{A:rtl} reports the setup and full results of the RTL simulations in Section~\ref{sec:complexity}.

\subsection{Polynomial evaluation via Horner's method}
\label{app:horner}

A polynomial of degree $N$ evaluated naively as $\sum_{k=0}^{N} a_{k} x^{k}$ requires $N(N+1)/2$ multiplications and $N$ additions. Horner's method~\citep{10.5555/579525} rewrites the polynomial in nested form,
\begin{equation}
  P(x) = a_{0} + x\bigl(a_{1} + x(a_{2} + \cdots + x\, a_{N})\cdots\bigr),
  \label{eq:horner}
\end{equation}
reducing the cost to exactly $N$ multiplications and $N$ additions, that is, $2N$ FLOPs. This count is FLOP-optimal for generic polynomial evaluation under the standard arithmetic model~\citep{pan1966methods}, and we adopt it as the basis for all polynomial cost estimates that follow.

\subsection{Exponential function as a unified primitive}
\label{app:exp}

We evaluate $\exp(x)$ via standard range reduction followed by a Maclaurin truncation~\citep{muller2018handbook}. The cost decomposes into three stages: a range reduction that maps the input into a bounded interval, a polynomial approximation evaluated on that interval, and a reconstruction step that recovers the original argument.

\paragraph{Range reduction.}
We decompose the input as
\begin{equation}
  x = k\ln 2 + r,
  \qquad k = \mathrm{round}\!\bigl(x \cdot \log_{2} e\bigr) \in \mathbb{Z},
  \qquad |r| \le \tfrac{\ln 2}{2} \approx 0.3466,
  \label{eq:exp-decomp}
\end{equation}
which gives the identity $e^{x} = 2^{k} \cdot e^{r}$. Since $\log_{2} e = 1/\ln 2$, the integer $k$ is the nearest integer to $x/\ln 2$, ensuring $|r| \le \ln 2/2$. The reduction itself costs four discrete operations: one multiplication ($x \cdot \log_{2} e$), one round-to-integer, one multiplication ($k \cdot \ln 2$), and one subtraction ($r = x - k\ln 2$).

\paragraph{Truncation bound.}
The Maclaurin truncation $E_{N}(r) = \sum_{k=0}^{N} r^{k}/k!$ has Lagrange remainder
\begin{equation}
  \bigl|e^{r} - E_{N}(r)\bigr|
  \;\le\; \frac{e^{|r|}\,|r|^{N+1}}{(N+1)!}
  \;\le\; \frac{\sqrt{2}\,|r|^{N+1}}{(N+1)!},
  \label{eq:exp-remainder}
\end{equation}
where the second inequality uses $e^{|r|} \le e^{\ln 2/2} = \sqrt{2}$. Evaluating~\eqref{eq:exp-remainder} at the boundary $|r| = \ln 2/2$ gives:
\[
  \begin{array}{c|cc}
    N & \text{degree} & \sqrt{2}\,(\ln 2/2)^{N+1}/(N+1)! \\
    \hline
    6 & 6 & 1.69 \times 10^{-7} \;>\; u \\
    7 & 7 & 7.30 \times 10^{-9} \;<\; u
  \end{array}
\]
Therefore $N = 7$ is the minimum polynomial order satisfying FP32 unit-roundoff accuracy, and Horner evaluation~\eqref{eq:horner} of degree $7$ costs $2 \cdot 7 = 14$ FLOPs.

\paragraph{Reconstruction.}
The factor $2^{k}$ is applied by adding $k$ to the FP32 exponent field, a single integer-arithmetic operation that we count as one FLOP.

\paragraph{Total cost.}
\begin{equation}
  \mathrm{FLOPs}_{\exp}
  \;=\; \underbrace{4}_{\text{range red.}}
        + \underbrace{14}_{\text{Horner }N=7}
        + \underbrace{1}_{\text{reconstr.}}
  \;=\; \mathbf{19}.
  \label{eq:exp-flops}
\end{equation}

\paragraph{Remark on production implementations.}
Numerous strategies exist for evaluating the exponential function in practice, ranging from straight Maclaurin or Chebyshev truncations to minimax (Remez) polynomials, table-driven hybrids that combine small look-up tables with low-degree polynomial corrections, and multi-phase schemes that fall back to extended precision for hard inputs~\citep{10.1145/63522.214389, muller2018handbook}. Each strategy trades polynomial degree, table size, and accuracy differently, and production libraries typically adopt a higher-degree minimax polynomial to obtain the required accuracy with a tight worst-case error bound. Our reported $N = 7$ is the \emph{mathematical} minimum derived from the Maclaurin truncation bound alone for FP32 unit-roundoff accuracy, and we adopt it as the FLOP baseline to obtain a tight lower bound on arithmetic cost.

\subsection{Transcendental functions via the unified \texorpdfstring{$\exp$}{exp} primitive}
\label{app:transcendentals}

The sigmoid and hyperbolic tangent are both expressed as direct compositions of a single $\exp$ evaluation, allowing the FLOP cost derived in Appendix~\ref{app:exp} to propagate to both. The error function needs a polynomial on top of the $\exp$ (Appendix~\ref{app:erf}). We treat $\exp$ as a unified primitive of $19$ FLOPs and account separately for the algebraic operations that compose it.

\subsubsection{Sigmoid}
\label{app:sigmoid}

The sigmoid is defined as $\sigma(x) = 1/(1 + e^{-x})$. Its evaluation decomposes into one negation, one exponential, one addition, and one division:
\begin{equation}
  \mathrm{FLOPs}_{\sigma}
  \;=\; \underbrace{1}_{-x}
        + \underbrace{19}_{\exp}
        + \underbrace{1}_{+1}
        + \underbrace{1}_{1/(\cdot)}
  \;=\; \mathbf{22}.
  \label{eq:sigmoid-flops}
\end{equation}

\subsubsection{Hyperbolic tangent}
\label{app:tanh}

Multiplying numerator and denominator of $\tanh(x) = (e^{x} - e^{-x}) / (e^{x} + e^{-x})$ by $e^{x}$ yields the algebraically equivalent form
\begin{equation}
  \tanh(x) = \frac{e^{2x} - 1}{e^{2x} + 1},
  \label{eq:tanh-decomp}
\end{equation}
which requires only a single $\exp$ evaluation. The cost decomposes into one doubling, one exponential, one subtraction, one addition, and one division:
\begin{equation}
  \mathrm{FLOPs}_{\tanh}
  \;=\; \underbrace{1}_{2x}
        + \underbrace{19}_{\exp}
        + \underbrace{1}_{-1}
        + \underbrace{1}_{+1}
        + \underbrace{1}_{\div}
  \;=\; \mathbf{23}.
  \label{eq:tanh-flops}
\end{equation}

\subsubsection{Error function}
\label{app:erf}

Unlike $\sigma$ and $\tanh$, the error function cannot be written in closed form in terms of $\exp$. We therefore use the approximation of~\citet{abramowitz1964handbook} (Eq.~7.1.26), which for $z \ge 0$ reads
\begin{equation}
  \mathrm{erf}(z) \approx 1 - t\,P(t)\,e^{-z^{2}}, \qquad t = \frac{1}{1 + p z}, \qquad P(t) = a_{1} + a_{2} t + a_{3} t^{2} + a_{4} t^{3} + a_{5} t^{4},
  \label{eq:erf-as}
\end{equation}
with $p = 0.3275911$ and five fixed coefficients $a_{1}, \dots, a_{5}$. Negative inputs follow from $\mathrm{erf}(-z) = -\mathrm{erf}(z)$. We do not count the absolute value or the sign restoration as FLOPs, because both only change the sign bit (as for \texttt{clip} in Appendix~\ref{app:gp-flops}). The cost decomposes into $t$ (one multiplication, one addition, and one division), $P(t)$ in Horner form~\eqref{eq:horner} of degree $4$, its product with $t$, one squaring, one negation, one exponential, one multiplication, and one subtraction:
\begin{equation}
  \mathrm{FLOPs}_{\mathrm{erf}}
  \;=\; \underbrace{3}_{t}
        + \underbrace{8}_{P(t)}
        + \underbrace{1}_{t \cdot P}
        + \underbrace{1}_{z^{2}}
        + \underbrace{1}_{-z^{2}}
        + \underbrace{19}_{\exp}
        + \underbrace{1}_{\times}
        + \underbrace{1}_{1-(\cdot)}
  \;=\; \mathbf{35}.
  \label{eq:erf-flops}
\end{equation}
The absolute error of~\eqref{eq:erf-as} is at most $1.5 \times 10^{-7}$, slightly above the FP32 unit roundoff $u$. An erf that is accurate to $u$ would need a longer polynomial, so $35$ FLOPs is a lower bound and, if anything, favours Derf.

\subsection{Per-token and per-layer FLOP cost of normalization methods}
\label{app:method-flops}

We now apply the FLOP convention established in Appendices~\ref{app:horner}--\ref{app:transcendentals} to the four normalization methods compared in this paper: standard LayerNorm, DyT, Derf, and our GP-evolved expressions. All counts exclude the per-channel affine transform $(w, b)$, which is shared across all methods and therefore does not affect the comparison.

\subsubsection{LayerNorm}
\label{app:ln-flops}

The per-token FLOP cost of LayerNorm follows directly from its arithmetic definition $y_{i} = (x_{i} - \mu)/\sqrt{\sigma^{2} + \epsilon}$, with $\mu = (1/d) \sum_{i} x_{i}$ and $\sigma^{2} = (1/d) \sum_{i} (x_{i} - \mu)^{2}$. Table~\ref{tab:layernorm_flops} enumerates the operations and the resulting FLOP count. The two reduction passes for $\mu$ and $\sigma^{2}$ couple all $d$ feature dimensions, creating the global cross-feature dependency that motivates the analysis in Section~\ref{sec:complexity}.

\begin{table}[h]
\centering
\caption{Per-token FLOP count for LayerNorm.}
\label{tab:layernorm_flops}
\begin{tabular}{@{}lll@{}}
\toprule
\textbf{Step}                   & \textbf{Operations}            & \textbf{FLOPs} \\
\midrule
$\sum_{i} x_{i}$                & $d - 1$ adds                   & $d - 1$ \\
$\mu = (\cdot)/d$               & $1$ div                        & $1$     \\
$x'_{i} = x_{i} - \mu$          & $d$ subs                       & $d$     \\
$(x'_{i})^{2}$                  & $d$ muls                       & $d$     \\
$\sum_{i} (\cdot)$              & $d - 1$ adds                   & $d - 1$ \\
$\sigma^{2} = (\cdot)/d$        & $1$ div                        & $1$     \\
$(\cdot) + \epsilon$            & $1$ add                        & $1$     \\
$\sqrt{(\cdot)}$                & $1$ sqrt                       & $1$     \\
$1/(\cdot)$                     & $1$ div                        & $1$     \\
$x'_{i} \cdot (\cdot)$          & $d$ muls                       & $d$     \\
\midrule
\textbf{Total}                  &                                & $\mathbf{5d + 3}$ \\
\bottomrule
\end{tabular}
\end{table}

\subsubsection{DyT}
\label{app:dyt-flops}

DyT is defined as $\mathrm{DyT}(x) = \tanh(\alpha x)$, with $\alpha$ a learnable scalar shared across the feature dimension. Each token therefore incurs $d$ multiplications for the input scaling and $d$ tanh evaluations, yielding $d + 23 d = 24d$ FLOPs per token under the unified $\exp$ primitive of Appendix~\ref{app:transcendentals}.

\subsubsection{Derf}
\label{app:derf-flops}

Derf is defined as $\mathrm{Derf}(x) = \mathrm{erf}(\alpha x + s)$, with learnable scalars $\alpha$ and $s$ shared across the feature dimension. Each token therefore incurs $d$ multiplications and $d$ additions for $\alpha x + s$, and $d$ erf evaluations, yielding $2d + 35d = 37d$ FLOPs per token (Appendix~\ref{app:erf}). Across all normalization layers, this is $925d$ in ViT-B and $1813d$ in ViT-L, and $272.9$ and $213.5$ MFLOPs per image in Swin-B and ConvNeXt-B (computed per image as described in Appendix~\ref{app:gp-flops}), about $1.5\times$ the cost of DyT.

\subsubsection{GP-evolved expressions}
\label{app:gp-flops}

The per-layer FLOP cost of our GP solutions is reported in the rightmost column of Tables~\ref{tab:gp_solutions_vitb}--\ref{tab:gp_solutions_convnextb} (Appendix~\ref{A:equations}), computed under the same convention. Each $\tanh$ contributes $23$ FLOPs and each $\sigma$ contributes $22$ FLOPs; \texttt{clip} operations are neglected because they reduce to a comparison and a select on hardware that supports conditional moves, neither of which we count as FLOPs. Aggregating across all normalization layers, the GP expressions in ViT-B ($25$ layers, $d = 768$) require $31d$ FLOPs per token, compared to $125d + 75$ for LayerNorm, $600d$ for DyT, and $925d$ for Derf. In ViT-L ($49$ layers, $d = 1024$), they require $85d$, compared to $245d + 147$, $1176d$, and $1813d$. Swin-B and ConvNeXt-B change their width $d$ between stages, so for these two we report the total cost per image instead (as in Figure~\ref{fig:complexity}, Right). We multiply the per-position cost of each layer by its width and by the number of positions it normalizes ($224 \times 224$ input), and sum over all layers. In Swin-B, the GP expressions require $16.1$ MFLOPs per image, compared to $37.0$ for LayerNorm, $177.0$ for DyT, and $272.9$ for Derf. In ConvNeXt-B, they require $8.4$ MFLOPs, compared to $28.9$, $138.5$, and $213.5$. A few layers in the first stage of Swin-B cost more than LayerNorm on their own (up to $6d$), but the total is still well below all three baselines.

We note one deliberate asymmetry between this deployment cost and the complexity metric used during the evolutionary search. Assigning \texttt{clip} its true cost of zero during the search makes redundant \texttt{clip} nesting complexity-neutral, and we observed the search exploiting this to stack such operations without any effect on the resulting function. We therefore charge \texttt{clip} a nominal cost of one FLOP during the search only, purely as a regularizer against this degenerate behaviour. All reported hardware costs, including those in Table~\ref{tab:gp_solutions_vitb} and Figure~\ref{fig:complexity}, use the true zero cost.

\subsection{Memory access analysis}
\label{app:memory}

Beyond arithmetic cost, normalization replacements differ in the volume of off-chip memory traffic they generate, which on bandwidth-bound hardware often dominates inference latency and energy. We quantify this traffic in bytes per token assuming FP32 storage ($4$ bytes per element) and a feature dimension $d$, counting only read traffic since the methods we compare differ structurally in how many times the input vector must be re-read from memory but not in their output write behaviour.

\paragraph{LayerNorm.}
The reduction operations for $\mu$ and $\sigma^{2}$ couple all $d$ feature dimensions, forcing two separate passes over the input vector. Pass 1 reads $x$ to compute the global statistics, and pass 2 reads $x$ again to apply the normalization. The cross-feature dependency in pass 1 forbids fusion with pass 2, since the per-element normalization requires $\mu$ and $\sigma^{2}$ to be available in full before any output element can be produced. The resulting read cost is
\begin{equation}
  \mathrm{Bytes}_{\mathrm{LN}}^{\mathrm{read}}
  \;=\; 2 \cdot 4d
  \;=\; 8d \;\;\text{bytes per token.}
\end{equation}
This count assumes that the vector does not fit in the buffer of the normalization unit, so that both passes read it from off-chip memory ($d > M$ in Section~\ref{sec:complexity}). When it fits, LayerNorm reads it once and makes its second pass on chip, at the cost of a buffer of at least $d$ values.

\paragraph{DyT, Derf, and GP.}
All three replacements are element-wise scalar mappings, requiring a single read of $x$ to produce each output element. The single-pass execution fuses naturally with adjacent operations and incurs only
\begin{equation}
  \mathrm{Bytes}_{\mathrm{DyT}}^{\mathrm{read}}
  \;=\; \mathrm{Bytes}_{\mathrm{Derf}}^{\mathrm{read}}
  \;=\; \mathrm{Bytes}_{\mathrm{GP}}^{\mathrm{read}}
  \;=\; 4d \;\;\text{bytes per token,}
\end{equation}
half the read traffic of LayerNorm when the vector does not fit in the buffer.

\paragraph{Aggregate.}
Across all $25$ normalization layers in ViT-B with sequence length $\ell = 197$ and feature dimension $d = 768$, the total read access per forward pass is $\mathrm{Bytes}_{\mathrm{LN}} = 25 \cdot \ell \cdot 8d \approx 29.0$~MB for LayerNorm and $\mathrm{Bytes}_{\mathrm{DyT}} = \mathrm{Bytes}_{\mathrm{GP}} = 25 \cdot \ell \cdot 4d \approx 14.5$~MB for DyT and GP.

\begin{figure}[tbp]
    \centering
    \begin{minipage}[b]{0.70\textwidth}
        \centering
        \raisebox{0pt}{\includegraphics[width=\linewidth]{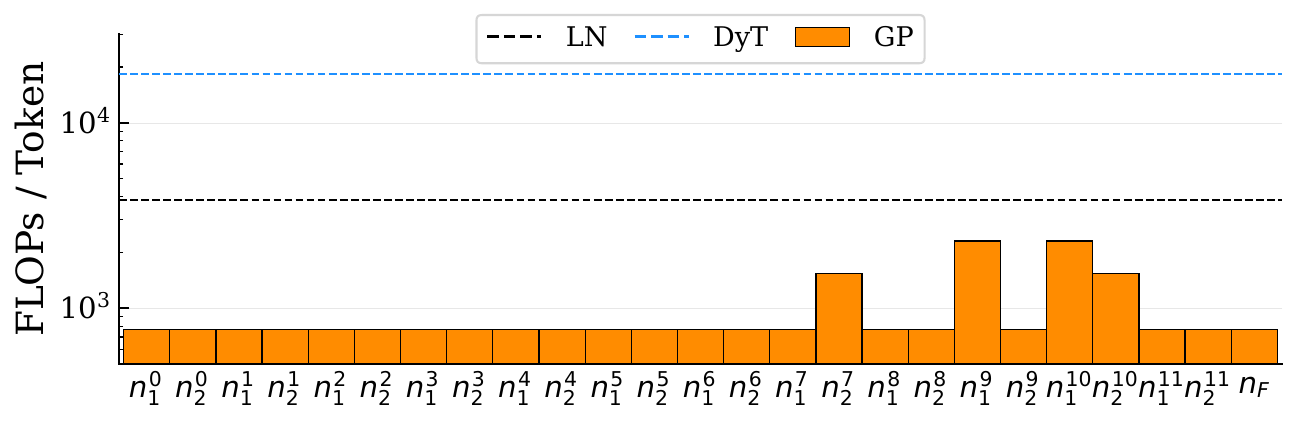}}
    \end{minipage}
    \hfill
    \begin{minipage}[b]{0.288\textwidth}
        \centering
        \raisebox{-6.9pt}{\includegraphics[width=\linewidth]{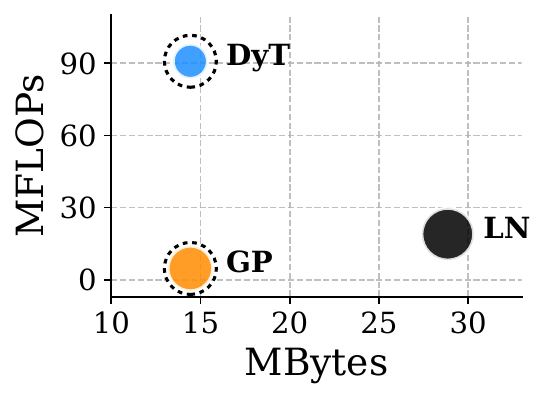}}
    \end{minipage}
\caption{\textbf{Computational and memory complexity of normalization replacements in ViT-B.}
\textbf{(Left)} Per-token FLOP counts for each of the 25 normalization layers, where $n_{j}^{i}$ denotes the $j$-th normalization within block $i$ and $n_{F}$ denotes the final normalization layer. Our hardware-aware GP search explicitly restricts FLOP usage, producing layer-specific expressions that are uniformly cheaper than both the DyT and LN baselines.
\textbf{(Right)} Aggregate complexity across all 25 normalization layers for a single forward pass (sequence length $\ell=197$, feature dimension $d=768$): total FLOPs (MFLOPs) versus total memory read access (MB). Bubble area is proportional to Top-1 accuracy; dashed circles around the GP and DyT markers show LN's bubble size.}
\label{fig:complexity}
\end{figure}

Figure~\ref{fig:complexity} summarizes the FLOP and memory counts for ViT-B, using the best variant of each method (LN, DyT-D, and GP-D, Table~\ref{tab:main_results}). Variants of one method differ only in parameter values, so they share the same hardware cost. Per layer (Left), every GP expression stays below both baselines, and the cost follows the mapping: the near-linear mappings in early blocks get single-FLOP expressions, while a few deeper layers with S-shaped responses need $2d$ or $3d$ FLOPs. In aggregate (Right), GP needs about $4\times$ fewer FLOPs than LayerNorm (seed 1) and DyT about $4.8\times$ more, at the read access computed above.

\subsection{RTL simulations}
\label{A:rtl}

Both chips are built on the Caravel SoC (Section~\ref{sec:complexity}). Its management SoC is a VexRiscv RV32I core with 1.5\,KB of internal SRAM and an external SPI flash that holds the program and data. Even in the optimistic case where the entire SRAM held vector data, which is not possible in practice since it also holds the stack and program data, it could store only 384 FP32 values. On the RISC-V core, the buffer is a region of this SRAM. The input vectors are stored in the flash in FP32. Each tile is read from the flash into the buffer and processed there, and the results are written out of the SoC over its Wishbone bus only for verification. On ASIC-NAU, the NAU is a pipelined datapath controlled by a wide VLIW-style instruction word, so a single instruction drives its shift, multiply, add, compare, and lookup table units in the same cycle while elements stream through it one at a time. The two vector sizes cover both regimes of Section~\ref{sec:complexity}: with $d = M$ the vector fits in one tile, and with $d = 2M$ LayerNorm must fetch every tile a second time. Because the energy excludes the external flash memory, the reported memory energy understates the true energy of every off-chip access.

Table~\ref{tab:rtl} gives the cycles and energy of both phases. At $d = 32$ the memory phase costs the same for all three methods, to within $0.5\%$ on both chips. At $d = 64$, LayerNorm's memory phase costs twice that of GP, while DyT and GP still read the same. In the compute phase, LayerNorm takes $3.2$--$3.3\times$ the cycles of GP and DyT $11$--$13\times$ on both chips. On the RISC-V core the compute energy ratios follow the cycle ratios, and on ASIC-NAU they are slightly higher ($3.9$--$4.1\times$ for LayerNorm and $13.1$--$13.4\times$ for DyT).

\begin{table}[t]
\centering
\caption{Cycles and energy per normalization layer from the RTL simulations. Ratios are relative to GP. Memory is reading the inputs from the flash, compute is the normalization. On ASIC-NAU the management SoC only moves the inputs into the NAU buffer, and all computation runs on the NAU. Energy covers the SoC (on ASIC-NAU including the NAU) and excludes the external flash memory. GP is averaged over the $25$ layers of ViT-B (seed 1).}
\label{tab:rtl}
\small
\setlength{\tabcolsep}{4pt}
\begin{tabular}{rl rr rr rr rr}
\toprule
 & & \multicolumn{2}{c}{Memory} & \multicolumn{2}{c}{Compute} & \multicolumn{4}{c}{Total} \\
\cmidrule(lr){3-4}\cmidrule(lr){5-6}\cmidrule(lr){7-10}
$d$ & Method & Cycles & $E$ ($\mu$J) & Cycles & $E$ ($\mu$J) & Cycles & $E$ ($\mu$J) & Speedup & $E$ ratio \\
\midrule
\multicolumn{10}{l}{\textit{RISC-V core}} \\
32 & LN  & 4{,}739 & 3.35 & 269{,}411     & 205.29  & 274{,}150     & 208.64  & 3.19$\times$  & 3.15$\times$ \\
   & DyT & 4{,}721 & 3.36 & 1{,}029{,}291 & 793.34  & 1{,}034{,}012 & 796.71  & 12.04$\times$ & 12.05$\times$ \\
   & GP  & 4{,}741 & 3.35 & 81{,}142      & 62.79   & 85{,}883      & 66.14   & 1.00$\times$  & 1.00$\times$ \\
\cmidrule(lr){1-10}
64 & LN  & 18{,}956 & 13.40 & 520{,}241     & 396.00  & 539{,}197     & 409.40  & 3.14$\times$  & 3.10$\times$ \\
   & DyT & 9{,}442  & 6.71  & 2{,}054{,}088 & 1582.90 & 2{,}063{,}530 & 1589.61 & 12.03$\times$ & 12.03$\times$ \\
   & GP  & 9{,}482  & 6.70  & 162{,}054     & 125.39  & 171{,}536     & 132.08  & 1.00$\times$  & 1.00$\times$ \\
\midrule
\multicolumn{10}{l}{\textit{ASIC-NAU}} \\
32 & LN  & 4{,}714 & 10.34 & 421     & 1.33 & 5{,}135  & 11.68 & 1.06$\times$ & 1.09$\times$ \\
   & DyT & 4{,}714 & 10.34 & 1{,}479 & 4.41 & 6{,}193  & 14.76 & 1.28$\times$ & 1.38$\times$ \\
   & GP  & 4{,}714 & 10.34 & 131     & 0.33 & 4{,}845  & 10.67 & 1.00$\times$ & 1.00$\times$ \\
\cmidrule(lr){1-10}
64 & LN  & 18{,}856 & 41.36 & 822     & 2.62 & 19{,}678 & 43.98 & 2.03$\times$ & 2.06$\times$ \\
   & DyT & 9{,}428  & 20.69 & 2{,}891 & 8.67 & 12{,}319 & 29.36 & 1.27$\times$ & 1.38$\times$ \\
   & GP  & 9{,}428  & 20.69 & 261     & 0.66 & 9{,}689  & 21.35 & 1.00$\times$ & 1.00$\times$ \\
\bottomrule
\end{tabular}
\end{table}

\newpage
\section{Choice of calibration images}
\label{A:calibration}
The mappings in Section~\ref{data_acq} come from a single batch of 64 ImageNet-1K validation images, and the 90/10 split is over the sampled points, so the held-out points come from the same images as the points used in the search. To test whether the evolved expressions depend on these images, we extract new ViT-B mappings from 64 randomly drawn ImageNet-1K training images and rerun the full GP search on them with the same settings (5 seeds). Table~\ref{tab:calibration} scores both sets of expressions on both sets of mappings. It also reports their Top-1 accuracy directly after replacement and after fine-tuning with distillation (GP-D, with the same hyperparameters as in Table~\ref{tab:main_results}).

Both sets fit the mappings they were not evolved on almost as well as their own (the MSE is at most $0.007$ higher). They also give the same accuracy at replacement ($51.1\%$ against $51.0\%$) and after fine-tuning ($84.29 \pm 0.05\%$ against $84.32 \pm 0.01\%$). This means that the choice of calibration images does not affect the fit or the final accuracy of the evolved expressions.

\begin{table}[htbp]
  \small
  \centering
  \caption{\textbf{Choice of calibration images (ViT-B).} MSE against the LayerNorm mappings from validation and training images, Top-1 accuracy (\%) directly after replacement (on a 10{,}000-image subset of the validation set), and peak Top-1 accuracy (\%) after 20 epochs of fine-tuning with distillation (mean $\pm$ standard deviation over the 5 evolved sets).}
  \label{tab:calibration}
  \begin{tabular}{lcccc}
    \toprule
     & \multicolumn{2}{c}{\textbf{MSE ($\downarrow$)}} & \multicolumn{2}{c}{\textbf{Top-1 ($\uparrow$)}} \\
    \cmidrule(lr){2-3} \cmidrule(lr){4-5}
    \textbf{Evolved on} & \textbf{Validation} & \textbf{Training} & \textbf{At replacement} & \textbf{GP-D} \\
    \midrule
    Validation images (main text) & $0.100 \pm 0.004$ & $0.107 \pm 0.003$ & $51.0 \pm 3.6$ & $84.32 \pm 0.01$ \\
    Training images & $0.099 \pm 0.003$ & $0.102 \pm 0.003$ & $51.1 \pm 3.2$ & $84.29 \pm 0.05$ \\
    \bottomrule
  \end{tabular}
\end{table}

\section{Ablations of the search settings}
\label{A:ablations}
The search in Section~\ref{symbolic_search} adds two settings to a plain MSE search: the tail penalty in the fitness function $\mathcal{F}$ and the FLOP-based cost objective. To measure what each of them does, we rerun the ViT-B search without it and keep all other settings the same. Table~\ref{tab:ablations} reports the fit, the cost, the number of sets that train, and the accuracy after fine-tuning with distillation (GP-D).

\textbf{No tail penalty.} Without the penalty ($\gamma = 0$), the search finds a closer fit (MSE $0.066$ against $0.100$), but $12$--$15$ of the $25$ expressions in each set grow linearly without bound outside the range of the data (against $1$--$5$ with the penalty). With the GP-D hyperparameters in float16, as in the rest of the paper, three of the five sets fail within the first two epochs with a NaN loss. We therefore reran all five sets in bfloat16, which has a much larger range than float16. None of these runs gives a NaN loss, but one set diverges to chance level after five epochs. The other four reach $78.77 \pm 4.94\%$, between $1.5$ and $11.8$ points below GP-D. This means that the penalty is needed for the evolved expressions to train reliably, even though it gives a higher MSE (as for the flexible clip family in Appendix~\ref{A:clip}).

\textbf{No cost objective.} We introduced the cost objective because a limit on the size of the tree alone does not keep the expressions cheap. To show this, we run the same search with the number of nodes in the tree (the standard complexity measure in GP) as the second objective instead of the estimated FLOPs, and select the best-fitting expression of each layer. All other search settings stay the same, and we tune the fine-tuning hyperparameters for these sets with the same grid search as for GP (Appendix~\ref{A:hyperparams}). The fit is slightly closer (MSE $0.085$ against $0.100$) and GP-D reaches about the same accuracy ($84.25 \pm 0.02\%$ against $84.32 \pm 0.01\%$). Even without the cost objective, $21$ of the $25$ expressions contain \texttt{clip} (seed 1), and only $4$ use $\tanh$ or $\sigma$ alone. The higher cost comes from the $\tanh$ and $\sigma$ terms in $11$ layers. However, the expressions cost $474d$ FLOPs per token over the $25$ layers, about $15\times$ more than GP ($31d$) and about $4\times$ more than LayerNorm ($125d + 75$).

\begin{table}[htbp]
  \small
  \centering
  \caption{\textbf{Ablations of the search settings (ViT-B).} MSE against the LayerNorm mappings, cost in FLOPs per token summed over the 25 layers (seed 1, Appendix~\ref{app:method-flops}), number of evolved sets that train, and peak Top-1 accuracy (\%) after 20 epochs of fine-tuning with distillation (mean $\pm$ standard deviation over the 5 sets). $^\dagger$Fine-tuned in bfloat16, since in float16 three of the five sets fail with a NaN loss.}
  \label{tab:ablations}
  \begin{tabular}{lcccc}
    \toprule
    \textbf{Search} & \textbf{MSE ($\downarrow$)} & \textbf{Cost ($\downarrow$)} & \textbf{Trains} & \textbf{Top-1 ($\uparrow$)} \\
    \midrule
    GP (Ours) & $0.100 \pm 0.004$ & $31d$ & 5/5 & $84.32 \pm 0.01$ \\
    No tail penalty & $0.066 \pm 0.003$ & $42d$ & 4/5$^\dagger$ & $64.10 \pm 33.08$ \\
    No cost objective & $0.085$ & $474d$ & 5/5 & $84.25 \pm 0.02$ \\
    \bottomrule
  \end{tabular}
\end{table}

\section{Fitting the clip shape found by the search}
\label{A:clip}
On all four architectures, every evolved expression contains at least one \texttt{clip} operation (Appendix~\ref{A:equations}). To measure how much of GP's result comes from this shape alone, and how much from searching the structure of each layer's expression, we remove the search and fit one fixed clip form directly to every layer of ViT-B, $f(x) = a \cdot \mathrm{clip}(bx)$ (the form the search selected for 21 of the 25 layers), with \texttt{clip} at $\pm 5.0$ as in the search. We fit $a$ and $b$ per layer on the same 50{,}000-point mappings, with the same fitness function $\mathcal{F}$ as the GP search (Section~\ref{symbolic_search}). As for GP, the constants stay fixed during re-alignment, and we fine-tune with the same hyperparameters as GP-D (Table~\ref{tab:appendix_hparams_vitb}). There is only one fitted set, so the five Clip-D runs vary the training seed, whereas the five GP-D runs use the five evolved sets.

Clip-D reaches a peak Top-1 accuracy of $84.22 \pm 0.03\%$, against $84.32 \pm 0.01\%$ for GP-D and $83.36 \pm 0.04\%$ for DyT-D (Table~\ref{tab:clip}). Fitting the shape found by the search therefore closes $0.86$ of the $0.96$ points between DyT-D and GP-D. The remaining gap is small but consistent: all five Clip-D runs are below the GP-D mean.

We also fitted a more flexible clip family, $a \cdot \mathrm{clip}(bx) + c + dx$, with all four constants fitted in every layer. It has the lowest MSE of all fitted functions ($0.055$, against $0.100$ for GP and $0.122$ for $a \cdot \mathrm{clip}(bx)$). However, the model keeps only about $5\%$ Top-1 accuracy at replacement and does not recover during fine-tuning (below $13\%$ after five epochs). The GP search, in contrast, adds terms such as $dx$ only in the layers where they help. So a lower MSE does not mean that a function works better inside the model.

\begin{table}[htbp]
  \small
  \centering
  \caption{\textbf{Fitting the clip shape found by the search (ViT-B).} MSE against the LayerNorm mappings and peak Top-1 accuracy (\%) after 20 epochs of fine-tuning with distillation (mean $\pm$ standard deviation over 5 runs). Clip-D fits $a \cdot \mathrm{clip}(bx)$ per layer with the GP fitness function and keeps the constants fixed. Clip-D varies the training seed, GP-D the evolved set.}
  \label{tab:clip}
  \begin{tabular}{lcc}
    \toprule
    \textbf{Method} & \textbf{MSE ($\downarrow$)} & \textbf{Top-1 ($\uparrow$)} \\
    \midrule
    DyT-D & $0.300$ & $83.36 \pm 0.04$ \\
    Clip-D & $0.122$ & $84.22 \pm 0.03$ \\
    GP-D (Ours) & $\mathbf{0.100 \pm 0.004}$ & $\mathbf{84.32 \pm 0.01}$ \\
    \bottomrule
  \end{tabular}
\end{table}

\end{document}